\pdfoutput=1 
\documentclass[final,5p,times,twocolumn]{elsarticle}

\usepackage{bm}
\usepackage{url}
\usepackage{amsmath,amssymb,amsfonts}
\usepackage{algorithm}
\usepackage{algorithmic} 
\usepackage{amsthm}
\usepackage{pifont}
\usepackage{hyperref}
\usepackage{soul}
\usepackage{graphicx}
\usepackage{float} 
\usepackage{upgreek}
\usepackage[caption=false]{subfig} 
\usepackage{textcomp}
\usepackage{xcolor}
\usepackage{pgfplots}
\usepackage{adjustbox}
\usepackage{multirow}
\usepackage{diagbox}
\usepackage{makecell}

\usepackage{booktabs}    
\usepackage{stfloats}    
\usepackage{microtype}   
\usepackage[normalem]{ulem} 
\usepackage{array}       
\usepackage{setspace}    
\usepackage{caption}
\makeatletter
\newenvironment{breakablealgorithm}
  {
   \begin{center}
     \refstepcounter{algorithm}
     \hrule height.8pt depth0pt \kern2pt
     \renewcommand{\caption}[2][\relax]{
       {\raggedright\textbf{\ALG@name~\thealgorithm} ##2\par}%
       \ifx\relax##1\relax
         \addcontentsline{loa}{algorithm}{\protect\numberline{\thealgorithm}##2}%
       \else
         \addcontentsline{loa}{algorithm}{\protect\numberline{\thealgorithm}##1}%
       \fi
       \kern2pt\hrule\kern2pt
     }
  }{
     \kern2pt\hrule\relax
   \end{center}
  }
\makeatother

\theoremstyle{plain}

\theoremstyle{definition}
\newtheorem{definition}{Definition}

\newtheorem{myrule}{Rule}

\theoremstyle{remark}

\newcommand{\tcblu}[1]{\textcolor{blue}{#1}}
\newcommand{\tcorg}[1]{\textcolor{orange}{#1}}

\begin{document}

\begin{frontmatter}

\title{Point Cloud Registration via Probabilistic Self-Update Local Correspondence and Line Vector Sets}

\author[1]{Kuo-Liang Chung\corref{cor1}}
\ead{klchung01@gmail.com}
\cortext[cor1]{Corresponding author}

\author[1]{Yu-Cheng Lin}
\author[1]{Wu-Chi Chen}

\affiliation[1]{
    organization={Department of Computer Science and Information Engineering, National Taiwan University of Science and Technology},
    city={Taipei},
    postcode={106335},
    country={Taiwan}
}

\begin{abstract}
Point cloud registration (PCR) is a fundamental task for integrating 3D observations in remote sensing applications. This paper proposes a fast and effective PCR algorithm utilizing probabilistic self-updating local correspondence and line vector sets. Our dual RANSAC interaction model comprises a global RANSAC evaluating the global correspondence set and a local RANSAC operating on dynamically updated local sets. Initially, these local sets are constructed using angle histogram statistics and line vector length preservation techniques. To improve accuracy, a probabilistic self-updating strategy refines the local sets after each interaction round. To reduce runtime, we introduce a global early termination condition that optimally balances accuracy and efficiency. Finally, a weighted singular value decomposition estimates the registration solution. Evaluations on public datasets demonstrate our algorithm achieves superior time efficiency and at least a 10\% root mean square error improvement over state-of-the-art methods. The C++ source code is publicly available at \href{https://github.com/ivpml84079/Probabilistic-Self-Update-Line-Vector-Set-Based-Point-Cloud-Registration}{https://github.com/ivpml84079/Probabilistic-Self-Update-Line-Vector-Set-Based-Point-Cloud-Registration}.
\end{abstract}

\begin{keyword}
Correspondence set \sep line vector set \sep point cloud registration \sep probabilistic self-update strategy \sep RANSAC interaction model
\end{keyword}

\end{frontmatter}

\section{Introduction}\label{sec:introduction}
This section is divided into three parts: the point cloud registration (PCR) problem, a brief review of representative previous works, and an outline of our research methods and main contributions.

\subsection{The Point Cloud Registration Problem}
PCR is a fundamental component in many remote sensing applications \citep{hong2020more, hu2023v2pnet, ghorbani2022novel, tao2023distinctive, li2020robust, huang2021robust, wang2022robust}, supporting the integration of 3D observations for environmental modeling, terrain reconstruction, and large-scale geospatial mapping.
Its objective is to determine a spatial transformation that aligns the source point cloud to the target point cloud. In this study, the spatial transformation includes both rotation and translation. Typically, the source and target point clouds are captured by LiDAR or RGB-D cameras at different times, positions, and viewpoints. Therefore, PCR remains a challenging task. In this paper, ``registration solution'' and ``transformation'' are used interchangeably.

Before registration, the initial correspondence set $\mathbf{C}$ is typically constructed using the fast point feature histograms (FPFH) method \citep{rusu2009fast} or the fully convolutional geometric feature (FCGF) method \citep{choy2019fully}, which is based on ResNet \citep{he2016deep} and UNet \citep{ronneberger2015u}. These methods compute feature descriptors for both source and target points and identify point pairs with similar descriptors as correspondences.

Existing PCR solutions primarily rely on iterative closest point (ICP) \citep{theiler2014keypoint} based methods and random sample consensus (RANSAC) \citep{fischler1981ransac} based methods. In this paper, based on the existing RANSAC interaction model which consists of a global RANSAC operating on the global correspondence set and a local RANSAC operating on the self-update local correspondence and line vector sets, we propose a new probabilistic self-update line vector set based PCR algorithm. Therefore, we briefly review ICP-based methods and focus primarily on RANSAC-based approaches. Readers are referred to the PCR survey papers \citep{LYU2024110408,xu2023point} for a comprehensive overview.

\subsection{The Representative Previous Works}

The ICP method was first proposed by Besl and McKay \citep{icp2}. At each iteration, ICP utilizes the quaternion method \citep{horn1987closed} to find a better spatial transformation by minimizing the distance between the aligned source points and the closest target points. However, ICP is highly sensitive to initialization; poor initial alignments often lead to reduced accuracy and increased time overhead. To address this, a principal component analysis (PCA) approach \citep{he2013method} was proposed to provide a more robust initialization. Furthermore, ICP is sensitive to outliers and low overlap, leading to slow convergence. To mitigate these issues, several robust variants have been developed, including the penalty function-based sparse ICP method \citep{bouaziz2013sparse}, the Anderson acceleration method \citep{pavlov2018aa}, and the robust symmetric point-to-plane distance based ICP \citep{LI2022219} method.

To accelerate the search for closest points in ICP, two strategies have been proposed: the angular-invariant feature approach \citep{JIANG20093839} and the curvature feature similarity approach \citep{yao2021point}. These methods enhance matching efficiency by reducing the search space at each iteration. To reduce computational costs, the 4-points congruent set (4PCS) method \citep{aiger20084} leverages subgraph isomorphism. This framework was further accelerated by super-4PCS \citep{mellado2014super} via smart indexing, and by keypoint-based PCS (K4PCS) \citep{theiler2014keypoint} using 3D difference-of-Gaussians and Harris corner detectors. Moreover, the fast global registration (FGR) method \citep{FGR} utilizes three rigid correspondences and optimizes the spatial transformation efficiently using the graduated nonconvexity (GNC) function \citep{LYU2024110408, xu2023point}. 

The Random Sample Consensus (RANSAC) method was first proposed by Fischler and Bolles \citep{fischler1981ransac} and has been widely used to solve the PCR problem due to its robustness and accuracy. RANSAC is an iterative method, and at each iteration, it first randomly samples some correspondences from the initial correspondence set $\mathbf{C}$ which is often constructed by using the fast point feature histograms (FPFH) method \citep{rusu2009fast} by comparing the similarity of the feature descriptors between the source points and the target points. Based on the sampled correspondences, a tentative registration solution is obtained, and then the inlier set and the confidence level are updated. If the specified confidence level, such as 99.5\%, is not reached, the above iteration is repeated until the confidence level exceeds the specified threshold. Although the accuracy of the resultant registration solution by RANSAC is robust, it is time consuming for low inlier rate cases. Experimental data demonstrated that for some testing point cloud datasets, FCGF+RANSAC is more accurate and faster than FPFH+RANSAC.

To improve the accuracy and speed of RANSAC, Chum \textit{et al.} \cite{chum2003locally} proposed a locally optimized RANSAC (LO-RANSAC) method by applying a local optimization technique to the registration solution estimated from the random sample. To alleviate the over-sampling problem in LO-RANSAC, Lebeda \textit{et al.} \cite{lebeda2012fixing} proposed a fixing the locally optimized RANSAC (FLO-RANSAC) method that modifies the iterative least squares by introducing a limit on the number of inliers used in the least squares computation. Barath and Matas \citep{barath2018graph} proposed a graph-cut based RANSAC method in which a graph-cut process is used to solve the registration solution by first employing the spatial coherence information of points and the residual factor into the quality evaluation of the model hypothesis. Next, by minimizing a derived energy function, the registration solution is obtained, achieving better registration accuracy and less execution time relative to the methods in \citep{fischler1981ransac}, \citep{torr2002bayesian}, \citep{raguram2012usac}, \citep{brachmann2019neural}. Recently, an improved variant of the graph-cut based RANSAC, called the GC-RANSAC \citep{barath2021gcransac}, was proposed by taking a new spatial coherence model, some components in the universal RANSAC \citep{barath2021gcransac}, and the scoring scheme in MAGSAC++ \citep{barath2020magsac++} into account.

Based on the given line vector set $\mathbf{L}$ which is constructed from $\mathbf{C}$, Yang and Carlone \cite{yang2019polynomial} proposed a truncated least squares estimation and semidefinite relaxation (TEASER) method. In TEASER, the scale parameter is first solved using the adaptive voting method on $\mathbf{L}$. Next, the three rotation parameters are solved using the semidefinite relaxation method, and then the three translation parameters are solved using the adaptive voting method. They provided an invariant proof to show that the PCR problem is decomposed into scale, rotation, and translation sub-problems. Later, Yang \textit{et al.} \cite{yang2020teaser} proposed an improved version of TEASER, called TEASER++, by using the graduated nonconvexity and the truncated least squares (GNC-TLS) to solve the three rotation parameters. Experimental data demonstrated that TEASER++ outperforms the methods in \citep{fischler1981ransac}, \citep{yang2015go}, \citep{yang2020graduated}, \citep{FGR}. 

To tackle high outlier rates, Shi \textit{et al.} \cite{shi2021robin} proposed a new method, called Reject Outliers Based on Invariants (ROBIN), using the maximum k-core concept to remove outliers. The accuracy of the ROBIN method is competitive with that of TEASER++. Based on the line vector set $\mathbf{L}$, to improve the topological graph model-based method \citep{li2020robust}, Li \textit{et al.} \cite{li2021onepoint} proposed a One-point RANSAC method. In their method, a local minimization strategy is used to refine the scale estimation iteratively until the scale with the largest consensus set is accepted as the optimal scale solution. Next, they applied the scale-annealing biweight function, differentiation technique, and weighted least squares method to solve the rotation parameters. Finally, they utilized the potential inliers to estimate the translation parameters. Their method is demonstrated to be superior to those in \citep{lebeda2012fixing}, \citep{theiler2014keypoint}, \citep{yang2020teaser}, \citep{cai2019practical}.

Differing from the ROBIN method, Sun \textit{et al.} \cite{sun2021ransic} proposed a random sampling with invariance and compatibility (RANSIC) method. In RANSIC, the theory of invariance and compatibility is utilized to identify inlier candidates from $\mathbf{C}$. The consensus set is then determined by comparing the compatibility between each new candidate, which comprises three correspondences and all the old candidates. This iterative random sampling process continues until a specified termination condition is met. RANSIC is quite fast due to the proposal of a centroid-based estimation for translation, but the translation error is obvious. Later, a centralized RANSAC-based point cloud registration (C-RANSAC) method \citep{wt2024cransac} was proposed. In C-RANSAC, a scale histogram-based approach is first proposed to construct a reduced line vector set $\mathbf{L}^{red}$ from $\mathbf{L}$. In addition, a local early termination condition was proposed to speed up C-RANSAC.

Motivated by the challenges of handling outliers in point cloud registration, under the second-order spatial compatibility ($\text{SC}^2$), Chen \textit{et al.} proposed an $\text{SC}^2$-PCR++ method \citep{chen2023sc2pcr}. In their derived probability model, the proposed $\text{SC}^2$ measurement assesses the global compatibility between correspondences. Base on the $\text{SC}^2$ matrix, their method involves calculating an eigenvector corresponding to the largest eigenvalue. Then, some seed correspondences are selected via the calculated eigenvector. For each seed correspondence, considering its neighbors, a seed-centered correspondence set is constructed, and then redundant sets are further removed. Finally, the final registration solution is selected from the candidates obtained by performing SVD on the retained seed-centered correspondence sets. Experimental results demonstrated that $\text{SC}^2$-PCR++ outperforms the methods in \citep{fischler1981ransac}, \citep{FGR}, \citep{barath2021gcransac}. 

Sun \cite{sun2024sucoft} proposed a robust PCR method based on a guaranteed supercore maximization technique and a flexible thresholding (SUCOFT) strategy. In SUCOFT, the $\text{SC}^2$ compatibility matrix is first built up, and then the purified correspondence set is obtained. Furthermore, a maximum supercore, which is not guaranteed to contain the maximum clique, is determined via an iterative, trimming-based fast K-supercore detection approach. Finally, singular value decomposition (SVD) is applied to the purified correspondence set to derive the final registration solution. 

In 2023, Han \textit{et al.} \cite{han2023grid} developed a grid graph-based point cloud registration (GGR) method. In the GGR method, the unorganized point cloud is first divided into a set of 3D grids. Next, a voting strategy based on feature descriptors is employed to measure the similarity between the source grid and target grid, which transforms the correspondence into a grid representation. Then, a graph matching strategy is utilized to capture spatial consistency from the grid-based correspondence set. Finally, the gird-based correspondences are hierarchically refined until accurate point-to-point correspondences are obtained. However, despite being sensitive to grid orientation, GGR does not require an accurate initial location estimation to operate effectively. Experimental data demonstrated that GGR achieves better registration recall and accuracy relative to RANSAC~\citep{fischler1981ransac}, RANSACo~\citep{zhou2018open3d}, and TEASER~\citep{yang2019polynomial}.

Zhang \textit{et al.} \cite{zhang20233d} employed the maximal cliques (MAC) into PCR. In MAC, a spatial compatibility graph is first constructed from initial correspondences, and then maximal cliques are extracted from the graph to generate registration hypotheses. However, MAC often suffers from the conditions with extremely severe outliers. To address the computational inefficiency of MAC, Li \textit{et al.} \cite{li2024robust} developed a maximal clique with adaptive voting (MCAV) method. Rather than extracting cliques from all matches in MAC, MCAV pre-filters initial correspondences to significantly accelerate the estimation process. In 2025, Zhang \textit{et al.} proposed an enhanced version, called MAC++~\citep{zhang2025mac++}. The two novelties of MAC++ are: (1) an improved hypothesis generation step utilizes putative seeds through voting to guide the construction of maximal clique pools, ensuring valid hypotheses are preserved, and (2) a progressive hypothesis evaluation step continuously reduces the solution space in a ``global clusters-cluster-individual'' manner. Finally, the transformation parameters are estimated from the surviving hypotheses, achieving better accuracy than MAC.

In 2024, Huang \textit{et al.} \cite{Huang2024tear} proposed a PCR method using truncated entry-wise absolute residuals (TEAR) to minimize the loss function. The TEAR method decomposes the PCR problem into two subproblems, each operating in a lower-dimensional space. Specifically, the rotation matrix is represented as three vectors. In the first step, the first rotation vector along with a corresponding translation parameter is estimated using a branch-and-bound approach, which prunes correspondences that do not contribute to the estimated solution. In the second step, the method estimates the second rotation vector along with a corresponding translation parameter, while ensuring that it is orthogonal to the vector obtained in the first step. Based on the estimations of the three vectors, the complete rotation matrix and translation vector are followed, and only the inlier correspondences are retained. Finally, SVD is applied to these retained correspondences to obtain the final registration solution. 

Currently, Xing \textit{et al.} \cite{xing2024efficient} proposed a single correspondence voting method, referred to as the SCVC method. In the SCVC method, based on the ratio test idea, the top $k$ ($k = 500$ in their experiment) distinctive correspondences are selected from $\mathbf{C}$. For each selected correspondence, the translation solution with three parameters is first estimated. Next, based on the normal vectors of the left and right feature points, two rotation parameters are solved. Then, the remaining rotation parameter is estimated using a Hough transform-based voting strategy. Finally, the registration solution is determined from the three candidate solutions. SCVC significantly improves registration accuracy in both outdoor and indoor scenes.

In 2025, Xu \textit{et al.} \cite{xu2025globally} developed a multi-agent globally spatial consistency (MAGSC) method to identify the maximum consensus set. In MAGSC, a graph is first constructed based on the distance consistency of initial correspondences. Representative nodes are then selected as seeds to form clusters by exploring their second-order neighbors, generating candidate consensus sets. Furthermore, outliers in the initial correspondence set $\mathbf{C}$ are removed based on the evaluation of high-order global spatial consistency. Finally, an adaptive loss function and a graduated nonconvexity strategy are applied to iteratively refine the registration parameters. However, the accuracy of MAGSC suffers from the cases when the number of inliers is extremely small; the runtime of MAGSC increases as the constructed graph becomes dense.

Using the transformer networks and the attention mechanism, Mohr \textit{et al.} \cite{mohr2025gafar} proposed a deep learning-based graph-attention feature augmentation for registration (GAFAR) method. In the GAFAR method, only sparse subsets of the source and target point clouds are utilized to build a light-weight and accurate registration pipeline. The key idea behind GAFAR is that the relevant information for successful registration is determined by the topology of the overlapping area between the source and target point clouds as well as the geometric relationships among points in the overlapping area at test time. Based on 3DMatch and KITTI datasets, two variants of GAFAR are trained, respectively. Although any variant achieves good accuracy performance when the testing point cloud pairs are from the same training dataset, it suffers from accuracy degradation when the testing point cloud pairs are different from the training datasets. Experimental data demonstrated better performance relative to the deep learning-based geometric transformer (GeoTransformer) method \citep{qin2022geometric}, in which the transformer framework is used to learn geometric features for robust superpoint registration.

\subsection{The Outline of Our Research Methods and Main Contributions}
\begin{figure*}[htbp] 
\captionsetup{justification=raggedright, singlelinecheck=false, labelsep=newline}
\centering 
\includegraphics[width=\textwidth]{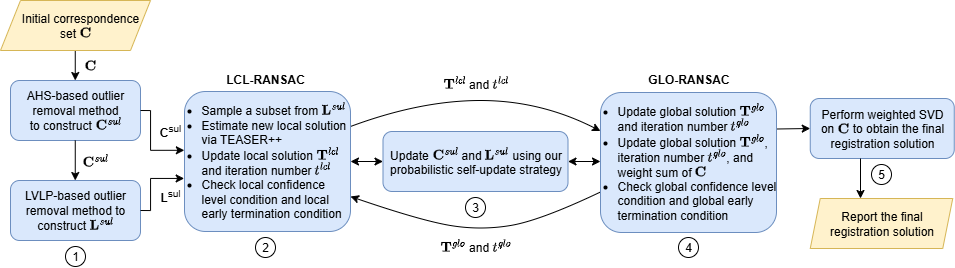} 
\caption{The pipelines of our PCR algorithm.}
\label{fig:pipeline}
\end{figure*}

In our study, the used RANSAC interaction model consists of one global RANSAC, which works on the global correspondence set $\mathbf{C}$, and one local RANSAC (LCL-RANSAC), which operates on the proposed self-update local (SUL) correspondence set $\mathbf{C}^{sul}$ and SUL line vector set $\mathbf{L}^{sul}$. The overall workflow of the proposed PCR algorithm consists of five pipelines, as illustrated in Figure~\ref{fig:pipeline}. The main and minor contributions of this study are summarized below.

\subsubsection{The main contribution}\label{sec:I-C1}
Different from existing outlier removal methods \citep{yang2019polynomial,yang2020teaser,li2021onepoint,sun2021ransic,wt2024cransac}, based on the initial correspondence set $\mathbf{C}$ and the newly calculated source and target normal vectors of each correspondence in $\mathbf{C}$, as depicted in the first pipeline of Figure~\ref{fig:pipeline}, we first propose a fast angle histogram- and statistic-based (AHS-based) and line vector length preservation-based (LVLP-based) outlier removal methods to construct the initial SUL correspondence and line vector sets, $\mathbf{C}^{sul}$ and $\mathbf{L}^{sul}$, respectively, which are used by LCL-RANSAC. However, both initial local sets may contain relatively more outliers. As depicted in the third pipeline of Figure~\ref{fig:pipeline}, after each RANSAC interaction round, the proposed probabilistic SUS is applied to enhance the compatibility of both local sets.

\subsubsection{The minor contribution}
In the previous RANSAC interaction model, three termination conditions are employed: the global confidence level condition, the local confidence level condition, and the local early termination condition \citep{wt2024cransac}. However, it still cannot reach a good trade-off between accuracy and runtime across varying inlier rates. To overcome this limitation, as depicted in the fourth pipeline of Figure~\ref{fig:pipeline}, we propose a global early termination condition defined by the maximal allowable number of RANSAC interaction rounds. This condition is integrated into the existing termination system to speed up runtime while maintaining good accuracy. When the global early termination condition, which is depicted in the second pipeline of Figure~\ref{fig:pipeline}, is reached, the weighted SVD technique is performed on the weighted correspondence set $\mathbf{C}$ to estimate the final registration solution, as depicted in the fifth pipeline.

To cover various scenarios, the three public testing datasets, 3DMatch \citep{zeng20173dmatch}, KITTI~\citep{geiger2012we}, and WHU-TLS \citep{WHU-TLS}, are used to evaluate both the accuracy and runtime of all considered methods. Seven accuracy metrics, rotation error, translation error, root mean square error (RMSE), median of square error (MeSE), precision, recall, and F1-score, and the required execution time in seconds are used to compare the performance of all considered methods. The comprehensive experimental data have demonstrated the accuracy and time merits of our algorithm relative to the six state-of-the-art methods: Graph-cut RANSAC \citep{barath2021gcransac}, TEASER++ \citep{yang2020teaser}, One-point RANSAC \citep{li2021onepoint}, RANSIC \citep{sun2021ransic}, $\text{SC}^2$-PCR++ \citep{chen2023sc2pcr}, and TEAR \citep{Huang2024tear}. In particular, our method achieves at least a 10\% improvement in RMSE compared with all comparative methods.

The remainder of this paper is organized as follows. In Section \ref{sec:sarlinevectorset}, the construction of the two initial local sets, $\mathbf{C}^{sul}$ and $\mathbf{L}^{sul}$, is presented. In Section \ref{sec:algorithm}, the proposed global early termination condition, probabilistic SUS, and pseudo code of our whole algorithm are presented. In Section \ref{sec:experiment}, comprehensive experimental results are presented to justify the accuracy and time merits of our algorithm. In Section \ref{sec:conclusions}, the conclusions are drawn.

\section{Construct the Initial Self-Update Local Correspondence and Line Vector Sets}\label{sec:sarlinevectorset} 

Given a source point cloud $P^s$ and a target point cloud $P^t$, initially the FCGF method \citep{choy2019fully} is applied to obtain the initial correspondence set $\mathbf{C}$ for the point cloud pair $(P^s, P^t)$. Different from existing methods, we first propose a fast AHS-based method to construct the initial SUL correspondence set $\mathbf{C}^{sul}$. Next, based on $\mathbf{C}^{sul}$, an LVLP-based method is proposed to construct the initial SUL line vector set $\mathbf{L}^{sul}$. As mentioned in Section~\ref{sec:I-C1}, the two constructed local sets, $\mathbf{C}^{sul}$ and $\mathbf{L}^{sul}$, may contain relatively more outliers. To enhance the compatibility of both local sets, after each RANSAC interaction round, a probabilistic SUS is proposed to update $\mathbf{C}^{sul}$ and $\mathbf{L}^{sul}$, as presented in Section~\ref{sec:saralgo}.

\subsection{Angle Histogram- and Statistic-Based (AHS-Based) Method to Construct the Initial SUL Correspondence Set}
\label{sec:csar}
Let the initial correspondence set be denoted as $\mathbf{C} = \{ \mathbf{c}_{i} \}_{i=1}^{N}$, where each correspondence $\mathbf{c}_{i} = (x_i, y_i)$ ($\in \mathbb{R}^3 \times \mathbb{R}^3$) satisfies $x_i \in P^{s}$ and $y_i \in P^{t}$. To construct the initial SUL correspondence set $\mathbf{C}^{sul}$, we first apply the subroutine ``NormalEstimation'' from the PCL library \citep{Rusu_ICRA2011_PCL}, which is based on the principal component analysis (PCA) technique, to estimate the normal vector of each feature point in $\mathbf{c}_i$ ($= (x_i, y_i)$). In the subroutine ``NormalEstimation'', covariance matrices of $x_i$ and $y_i$ are constructed using the 20 nearest neighbors of each feature point. The eigenvector corresponding to the smallest eigenvalue of the covariance matrix is used as the normal vector of the feature point. Let the normal vectors of $x_i$ and $y_i$ be denoted as $\mathbf{N}_{x_i}$ and $\mathbf{N}_{y_i}$, respectively. For fairness, the runtime required in obtaining the two normal vectors of each correspondence is included in the total runtime of our algorithm.

\begin{figure*}[htbp] 
\captionsetup{labelsep=period}
\centering
\begin{minipage}[t]{0.3\textwidth}
    \centering
    \includegraphics[width=\textwidth]{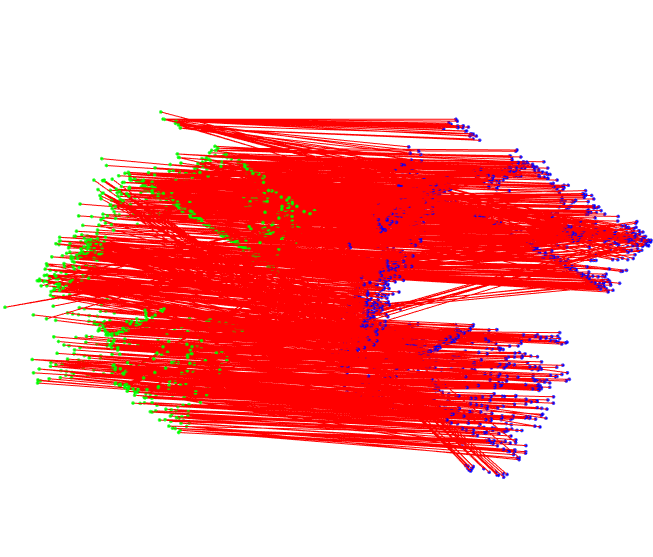}
    \caption*{(a)}
\end{minipage}\hfill
\begin{minipage}[t]{0.3\textwidth}
    \centering
    \includegraphics[width=\textwidth]{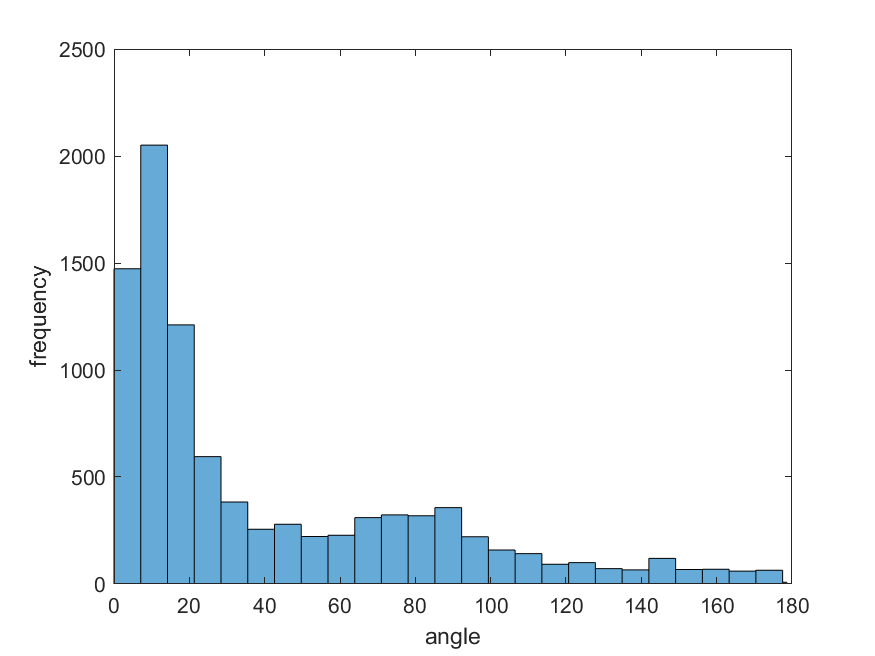}
    \caption*{(b)}
\end{minipage}\hfill
\begin{minipage}[t]{0.3\textwidth}
    \centering
    \includegraphics[width=\textwidth]{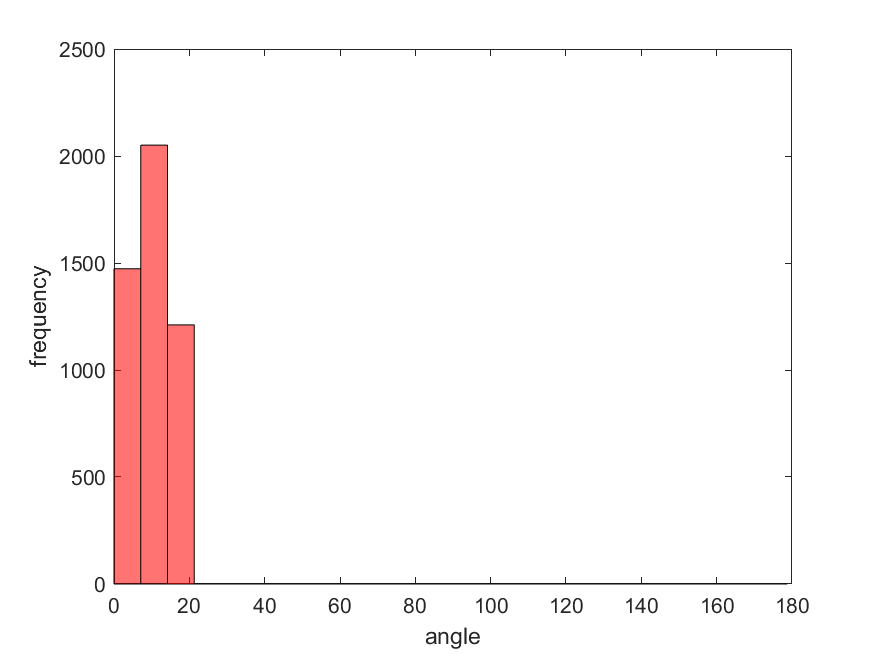}
    \caption*{(c)}
\end{minipage}

\vspace{10pt}

\begin{minipage}[t]{0.3\textwidth}
    \centering
    \includegraphics[width=\textwidth]{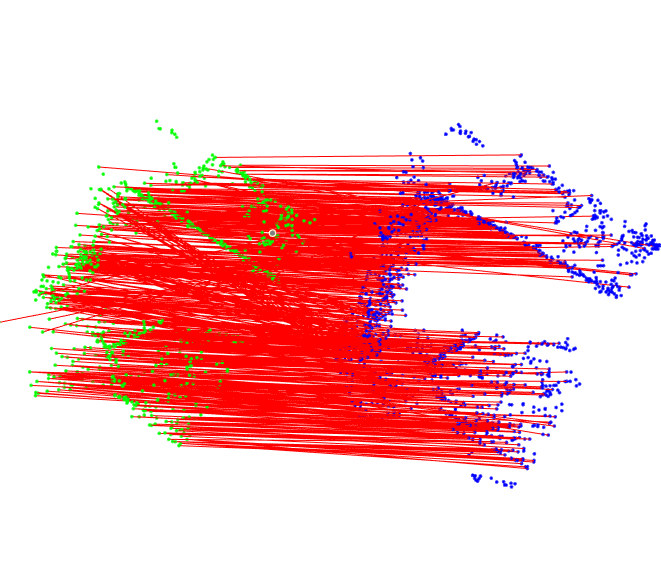}
    \caption*{(d)}
\end{minipage}\hfill
\begin{minipage}[t]{0.3\textwidth}
    \centering
    \includegraphics[width=\textwidth]{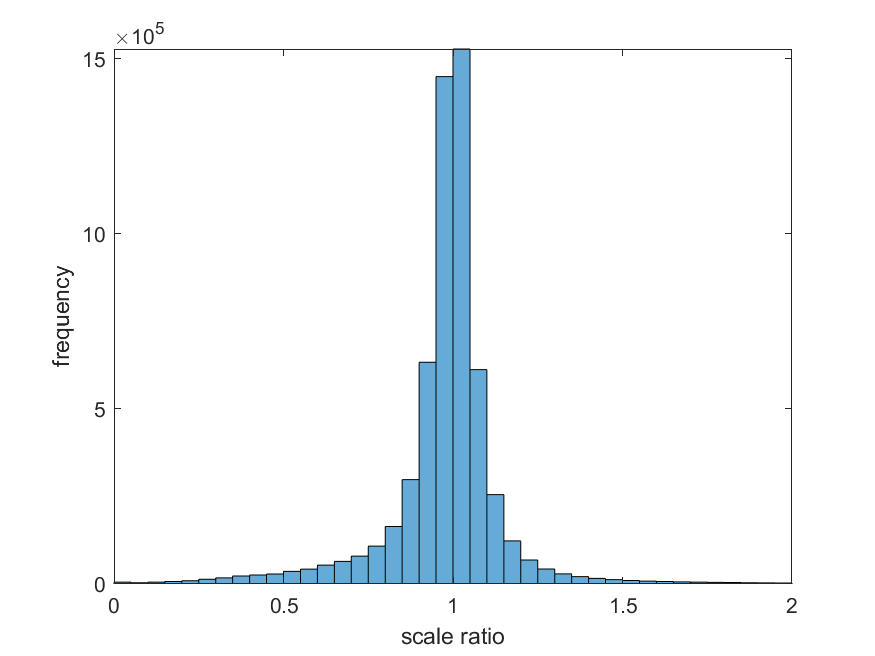}
    \caption*{(e)}
\end{minipage}\hfill
\begin{minipage}[t]{0.3\textwidth}
    \centering
    \includegraphics[width=\textwidth]{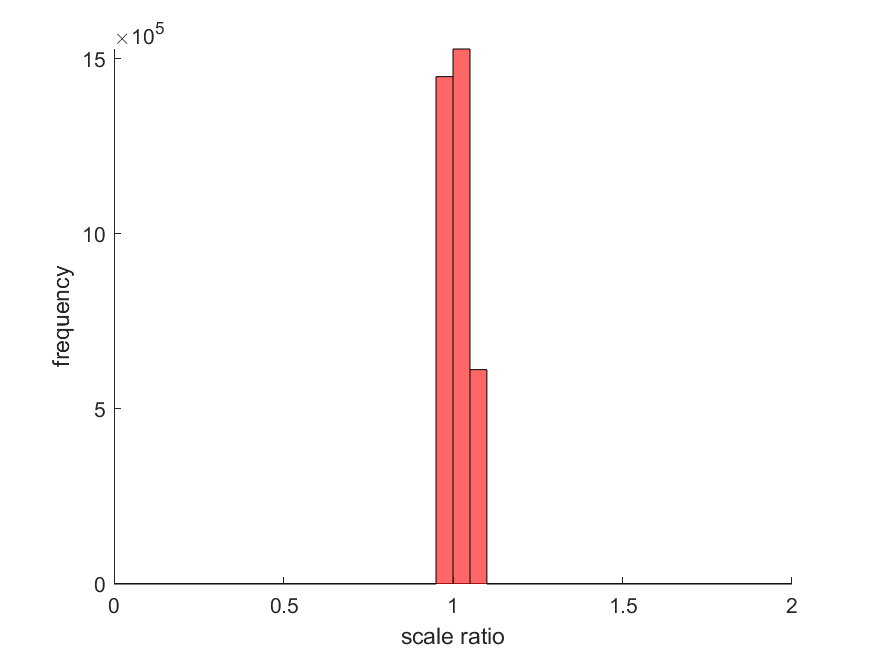}
    \caption*{(f)}
\end{minipage}

\caption{The construction of the initial SUL correspondence and line vector sets using the proposed AHS-based and LVLP-based methods. (a) The point cloud pair in which the source and target points are marked in green and blue, respectively, and the correspondences in $\mathbf{C}$ are denoted by red lines. (b) The angle histogram $\mathbf{H}(\theta)$. (c) The AHS-based angle histogram $\mathbf{H'}(\theta)$. (d) The initial SUL correspondence set $\mathbf{C}^{sul}$. (e) The scale ratio histogram $\mathbf{H}^{sr}$ of $\mathbf{C}^{sul}$. (f) The LVLP-based scale ratio histogram $\mathbf{H}'^{sr}$ of the initial SUL line vector set $\mathbf{L}^{sul}$.}
\label{fig:PCR_example}
\end{figure*} 

The angle between $\mathbf{N}_{x_i}$ and $\mathbf{N}_{y_i}$ is expressed as
\begin{equation}
\theta(\mathbf{c}_{i}) = \cos^{-1} \left( \mathbf{N}_{x_i}^{\text{norm}} \cdot \mathbf{N}_{y_i}^{\text{norm}} \right)
\label{eq:thetac}
\end{equation}
with $\theta \in [0, \pi]$, $\mathbf{N}_{x_i}^{\text{norm}} =  \frac{\mathbf{N}_{x_i}}{\|\mathbf{N}_{x_i}\|}$, and $\mathbf{N}_{y_i}^{\text{norm}} =  \frac{\mathbf{N}_{y_i}}{\|\mathbf{N}_{y_i}\|}$. Based on the calculated angles of all correspondences in $\mathbf{C}$, the angle histogram $\mathbf{H}(\theta)$ is constructed, where x-axis denotes the angle parameter $\theta$ and the y-axis denotes the frequency.

To determine the proper bin width of $\mathbf{H}(\theta)$, using Scott's rule \citep{scott1979scottrule}, the bin width $w$ is expressed as
\begin{equation}
\begin{aligned}
w = \frac{3.49 \cdot \sigma}{\sqrt[3]{|\mathbf{C}|}}
\end{aligned}
\label{eq:scott}
\end{equation}
\noindent with
\begin{equation}\label{eq:sigma}
    \sigma = \sqrt{\frac{1}{|\mathbf{C}|} \sum_{i=1}^{|\mathbf{C}|} (\theta(\mathbf{c}_i) - \mu)^2}
\end{equation}
\begin{equation}
    \mu = \frac{1}{|\mathbf{C}|} \sum_{i=1}^{|\mathbf{C}|} \theta(\mathbf{c}_i)
\end{equation}
\noindent where $\mu$ and $\sigma$ denote the mean and standard deviation of the angle distribution in $\mathbf{H}(\theta)$. Because the angle $\theta$ is within $[0, \pi)$, by Eq. (\ref{eq:scott}), the number of bins in $\mathbf{H}(\theta)$ is given by $N = \lceil\frac{\pi}{w}\rceil$. The first bin corresponds to the range $[0, w)$, and the $i$-th bin corresponds to $[(i - 1) \cdot w, i \cdot w)$. Scott’s rule aims to minimize the mean integrated squared error (MISE), thereby balancing between data spread and sample size and achieving some accuracy and efficiency benefits in the subsequent construction for $\mathbf{C}^{sul}$ and $\mathbf{L}^{sul}$. 

Based on the constructed angle histogram $\mathbf{H}(\theta)$, an AHS-based method is proposed to construct the initial SUL correspondence set $\mathbf{C}^{sul}$ by removing a certain portion of outliers. Inspired by the statistical analysis \citep{freund2003statistical}, $\mathbf{C}^{sul}$ is constructed by collecting the correspondences from all qualified bins in $\mathbf{H}(\theta)$, where each qualified bin means the frequency of the collected correspondences in the qualified bin is greater than the threshold $T_f$. In the third paragraph of Section \ref{sec:parameter}, the choice of $T_f$ is discussed in detail. For convenience, the histogram consisting of only the qualified bins is denoted by $\mathbf{H'}(\theta)$. In Figure~\ref{fig:PCR_example}(a), the initial correspondence set $\mathbf{C}$ is denoted as a set of red lines; the source and target point clouds are denoted by green and blue points, respectively. Figure~\ref{fig:PCR_example}(b) illustrates the constructed angle histogram $\mathbf{H}(\theta)$. After performing the AHS-based method on Figure~\ref{fig:PCR_example}(b), the constructed histogram $\mathbf{H'}(\theta)$ and the corresponding initial SUL correspondence set $\mathbf{C}^{sul}$ are depicted in Figure~\ref{fig:PCR_example}(c) and Figure~\ref{fig:PCR_example}(d), respectively. In this case, the improvement ratio of the number of correspondences in $\mathbf{H'}(\theta)$ over that in $\mathbf{H}(\theta)$ is 62.93\% (= $\frac{9248 - 3428}{9248}$).

\subsection{Line Vector Length Preservation-Based (LVLP-Based) Method to Construct the Initial SUL Line Vector Set} \label{sec:sul}
To construct the initial SUL line vector set $\mathbf{L}^{sul}$, based on $\mathbf{C}^{sul}$, a tentative line vector set $\mathbf{L}' = \{(v_{x}^{i,j}, v_{y}^{i,j}) \mid 1 \leq i < j \leq |\mathbf{C}^{sul}|\}$ is first constructed, where $v_{x}^{i,j} = x_i - x_j$, $v_{y}^{i,j} = y_i - y_j$, and $|\mathbf{L}'| = \frac{|\mathbf{C}^{sul}|(|\mathbf{C}^{sul}| - 1)}{2}$. For each line vector $\mathbf{l}_{i,j} = (v_{x}^{i,j}, v_{y}^{i,j}) \in \mathbf{L}'$, the scale ratio is calculated by $s_{i,j} = \frac{||v_{x}^{i,j}||}{||v_{y}^{i,j}||}$. Subsequently, as illustrated in Figure~\ref{fig:PCR_example}(e), the scale ratio histogram $\mathbf{H}^{sr}$ is constructed. Here, the computational complexity for constructing $\mathbf{H}^{sr}$ is $\mathcal{O}\!\left(|\mathbf{C}^{sul}|^2\right)$, where Big-O notation denotes an upper bound time complexity \citep{cormen2022introduction}.

To filter out unreliable line vectors in $\mathbf{L}'$, the LVLP-based approach only collects those line vectors located in the highest bin of $\mathbf{H}^{sr}$ and its immediate left and right neighboring bins. The scale ratios of the three collected bins of $\mathbf{H}^{sr}$ are close to 1, indicating an LVLP effect. Figure~\ref{fig:PCR_example}(f) illustrates the LVLP-based scale ratio histogram $\mathbf{H}'^{sr}$, in which the collected line vectors in $\mathbf{H}'^{sr}$ serve as the initial SUL line vector set $\mathbf{L}^{sul}$. Because the AHS-based outlier removal approach considers the angular similarity between the normal vectors of the source and target feature points for each correspondence in $\mathbf{C}$, the constructed local correspondence set $\mathbf{C}^{sul}$ is more reliable than that constructed using the scale histogram-based approach in \citep{wt2024cransac}. Consequently, a more reliable local line vector set $\mathbf{L}^{sul}$ is obtained.

\section{The Proposed Global Early Termination, Self-Update Strategy, and PCR Algorithm} \label{sec:algorithm}
We first present how to combine the proposed global early termination condition with three existing termination conditions, namely the global confidence level condition, the local confidence level condition, and the local early termination condition, to constitute a complete termination system used by our PCR algorithm, achieving a good trade-off between registration accuracy and time efficiency. Next, we introduce the proposed SUS, which iteratively updates the SUL correspondence and line vector sets, $\mathbf{C}^{sul}$ and $\mathbf{L}^{sul}$, to gradually enhance their compatibility and increase the registration accuracy. Finally, the pseudo code of our whole PCR algorithm is presented.

\subsection{Combine Our Global Early Termination Condition with Three Existing Termination Conditions}
As illustrated in the second pipeline of Figure~\ref{fig:pipeline}, the local confidence level condition and the local early termination condition constitute the local termination system for LCL-RANSAC. As depicted in the fourth pipeline of Figure~\ref{fig:pipeline}, the global confidence level condition and the global early termination condition constitute the global termination system for GLO-RANSAC. As a result, the four termination conditions constitute a complete termination system for our PCR algorithm.

\subsubsection{Two local termination conditions used in LCL-RANSAC}\label{sec:lclterminate}
Once LCL-RANSAC receives the current best global registration solution $\mathbf{T}^{glo}$ from GLO-RANSAC, where $\mathbf{T}^{glo}$ consists of a global rotation solution $R^{glo}$ and a global translation solution $T^{glo}$, LCL-RANSAC first creates a line vector subset $\mathbf{L}^{sul}_{sub}$ by randomly selecting $\alpha$\% of the line vectors from $\mathbf{L}^{sul}$. The choice of $\alpha$ is discussed in Section~\ref{sec:parameter}. Based on $\mathbf{L}^{sul}_{sub}$, LCL-RANSAC constructs a basic line vector set $\mathbf{L}^{sul}_{basic}$ by randomly selecting $\beta$\% of the line vectors from $\mathbf{L}^{sul}_{sub}$, and then performs the TEASER++ method~\citep{yang2020teaser} on $\mathbf{L}^{sul}_{sub}$ to estimate a new local registration solution, i.e., a new local transformation. The choice of $\beta$ is discussed in Section~\ref{sec:parameter}. Then, LCL-RANSAC updates the current best transformation $\mathbf{T}^{lcl}$. The local iteration number, saved in $t^{lcl}$, is increased by 1.

For LCL-RANSAC, the local early termination condition \citep{wt2024cransac} holds when the current local transformation and the received global transformation satisfy the following two inequalities:
\begin{align} \label{eq:inherit}
\arccos{\left(\frac{\text{tr}\big(R^{glo}(R^{lcl})^{\text{t}}\big)-1}{2}\right)} &\leq 0.01 \notag \\
\|T^{glo} - T^{lcl}\| \leq \tau
\end{align}
where the symbol ``t'' denotes the transpose operation; following the same parameter setting \citep{yang2020teaser}, the noise bound $\tau$ is set to 0.05.

When Equation~(\ref{eq:inherit}) holds, i.e., when the local early termination condition holds, it indicates that the estimated local transformation $\mathbf{T}^{lcl}$ is close to the received global transformation $\mathbf{T}^{glo}$. This is equivalent to estimating $\mathbf{T}^{lcl}$ by running the global hypothesis-and-verification (HAV) process for $t^{glo} + t^{lcl}$ iterations, where $t^{glo}$ denotes the number of iterations already spent in GLO-RANSAC. At this moment, instead of waiting until the local confidence level condition 99.5\% holds, LCL-RANSAC is immediately terminated, and according to the iteration number inherit rule, LCL-RANSAC directly sends back the accumulated iteration number, $t^{glo} + t^{lcl}$, and the estimated local transformation $\mathbf{T}^{lcl}$ to GLO-RANSAC.

If the local early termination condition in Equation~(\ref{eq:inherit}) does not hold, i.e., $\mathbf{T}^{lcl}$ is not sufficiently close to $\mathbf{T}^{glo}$, LCL-RANSAC further checks whether the local confidence level condition holds. If the local confidence level $\mathbf{Cl}^{lcl}$ $(= 1 - (1 - |{Ir}^{lcl}|)^{t^{lcl}})$, where $|{Ir}^{lcl}|$ denotes the currently largest local inlier rate, is larger than or equal to 99.5\%, LCL-RANSAC returns $\mathbf{T}^{lcl}$ and $t^{lcl}$ to GLO-RANSAC; otherwise, LCL-RANSAC estimates the next transformation and updates the currently best transformation $\mathbf{T}^{lcl}$. By the same argument, LCL-RANSAC repeats the above local transformation estimation and termination condition checking process until one of the two local termination conditions holds. When one local termination condition holds, the local iteration number $t^{lcl}$ is updated by
\begin{align}\label{eq:tlcl_update}
t^{lcl} :=
\begin{cases} 
t^{glo} + t^{lcl} &\text{if local early termination condition holds} \\
t^{lcl} &\text{if local confidence level condition holds}
\end{cases}
\end{align}

In Equation~(\ref{eq:tlcl_update}), the first assignment operation ``$t^{lcl} := t^{glo} + t^{lcl}$'' means that when the local early termination condition (see Equation~(\ref{eq:inherit})) holds, the new local iteration number equals the sum of the current local and global iteration number, thereby achieving a global iteration number inherit effect.

\subsubsection{Two global termination conditions used in GLO-RANSAC}
Once GLO-RANSAC receives the local transformation $\mathbf{T}^{lcl}$ and the local iteration number $t^{lcl}$, GLO-RANSAC updates the global transformation $\mathbf{T}^{glo}$, the inlier set $Ir^{glo}$, the iteration number $t^{glo}$ $(= t^{glo} + t^{lcl})$, and the confidence level $\mathbf{Cl}^{glo}$ $(= 1 - (1 - |{Ir}^{glo}|)^{t^{glo}})$. The traditional global confidence level condition and the specified maximal RANSAC interaction rounds constitute the termination system for GLO-RANSAC.

The maximal number of allowable RANSAC interaction rounds, denoted by $R_{max}$, is a specified parameter. Section~\ref{sec:parameter} discusses how to determine a good choice of the parameter $R_{max}$ as our global early termination condition. Let $R$ denote the current RANSAC interaction rounds spent. The RANSAC interaction model begins with GLO-RANSAC, and the initial value of $R$ is set to $0$. For $R = 0$, GLO-RANSAC sends the initial global solution and iteration number to LCL-RANSAC. After LCL-RANSAC completes its local registration task, the value of $R$ is increased by $1$.

If $\mathbf{Cl}^{glo} < 99.5\%$ and $R < R_{max}$, the proposed probabilistic SUS, which will be detailed in Section~\ref{sec:saralgo}, is applied to update the local correspondence and line vector sets. Next, GLO-RANSAC utilizes the received local registration solution and iteration number to continue the next interaction round. Otherwise, if $99.5\% \leq \mathbf{Cl}^{glo}$ or $R = R_{max}$, our algorithm is terminated and reports the final registration solution.

\subsection{The Proposed Probabilistic Self-Update Strategy for Updating Local Correspondence and Line Vector Sets}\label{sec:saralgo}

As mentioned in Section~\ref{sec:csar}, in the initial SUL correspondence and line vector sets, to a large extent, outliers have been removed, but to a certain extent, some outliers may still remain. Conversely, some other potential inliers may be discarded when constructing the two local sets. To address this limitation, after performing one RANSAC interaction round, we propose a probabilistic SUS to update the local correspondence set $\mathbf{C}^{sul}$. In this strategy, some potential global correspondences from $\mathbf{C}$ are added to $\mathbf{C}^{sul}$, while some unreliable local correspondences from $\mathbf{C}^{sul}$ are removed, thereby gradually enhancing the compatibility of $\mathbf{C}^{sul}$. Once $\mathbf{C}^{sul}$ is updated, each included global (or removed local) correspondence is used to assist updating the local line vector set $\mathbf{L}^{sul}$ quickly. The ablation study in Section~\ref{sec:ablation study} confirms the registration accuracy improvement effect of the proposed probabilistic SUS.

\subsubsection{Determine whether a potential global correspondence $\mathbf{c}_i$ can be truly included in $\mathbf{C}^{sul}$ or not}

\begin{figure}[htbp] 
    \centering
    \includegraphics[width=\linewidth]{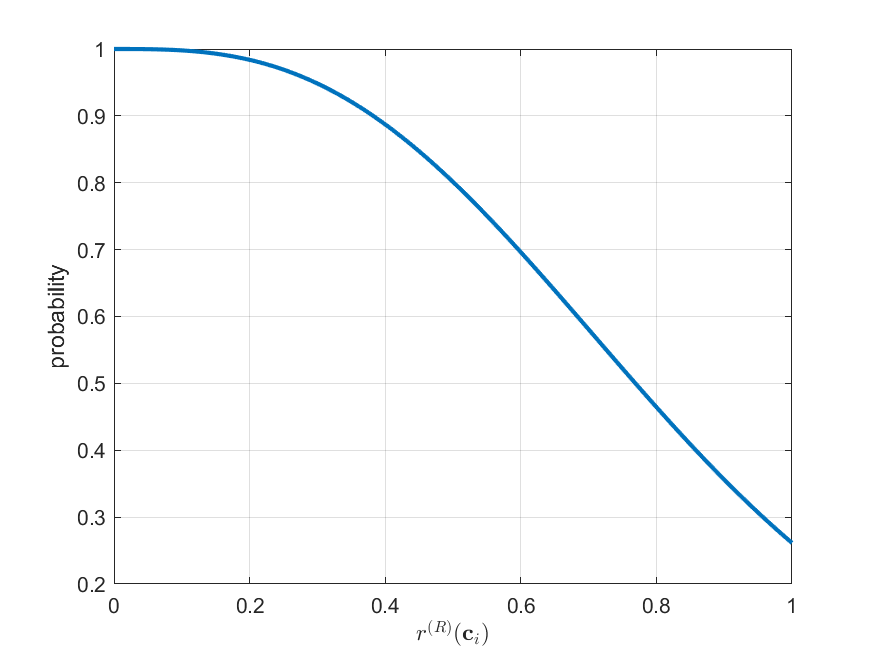} 
    \caption{The true inlier probability of the potential global inlier $\mathbf{c}_i$ according to Equation~(\ref{eq:probability}).}
    \label{fig:gamma_fig}
\end{figure}

After GLO-RANSAC updates $\mathbf{T}^{glo}$, ${Ir}^{glo}$, $t^{glo}$, and $\mathbf{Cl}^{glo}$ at the end of each RANSAC interaction round, if the condition ``$\mathbf{Cl}^{glo} < 99.5\%$ and $R < 5$'' holds, the weight sum of each global correspondence in ${Ir}^{glo}$ is incremented by one. In our algorithm, a potential global correspondence $\mathbf{c}_i$ in $\mathbf{C}$ is defined below.

\begin{definition}\label{def:D1}\sl{One global correspondence $\mathbf{c}_i$ in the current global inlier set ${Ir}^{glo}$ but not in the local SUL correspondence set $\mathbf{C}^{sul}$, i.e., $\mathbf{c}_i \in {Ir}^{glo} \cap (\mathbf{C} \backslash \mathbf{C}^{sul})$, is a potential global inlier.}
\end{definition}

Under the specified residual threshold $T_r$, whose value selection is discussed in the first paragraph of Section~\ref{sec:parameter}, the residual of a potential global inlier $\mathbf{c}_i$ at the $R$-th interaction round, denoted by $r^{(R)}(\mathbf{c}_i)$, is less than $T_r$ according to Definition~\ref{def:D1}. In what follows, we propose the first rule that a potential global inlier $\mathbf{c}_i$ can be included in the local SUL correspondence set $\mathbf{C}^{sul}$.

\begin{myrule}\label{rule:R1}\sl{For a potential global inlier $\mathbf{c}_i$ at the $R$-th interaction round, if at the $(R - 1)$-th round, the residual of $\mathbf{c}_i$, denoted by $r^{(R-1)}(\mathbf{c}_i)$, is also less than $T_r$, it indicates that $\mathbf{c}_i$ was also a potential global inlier at the $(R - 1)$-th round. For this case, i.e.,  $r^{(R-1)}(\mathbf{c}_i) < T_r$ and $r^{(R)}(\mathbf{c}_i) < T_r$, following the probability discussion in \citep{barath2020magsac++}, the true inlier probability of $\mathbf{c}_i$ is 1, we thus include the potential global correspondence $\mathbf{c}_i$ in $\mathbf{C}^{sul}$.}
\end{myrule}

If at the $(R - 1)$-th round, the global correspondence $\mathbf{c}_i$ is identified as an outlier, i.e., $r^{(R-1)}(\mathbf{c}_i) \geq T_r$, it yields $r^{(R-1)}(\mathbf{c}_i)\geq T_r$ and $r^{(R)}(\mathbf{c}_i) < T_r$. For this case, $\mathbf{c}_i$ is an inlier at the current round, but is an outlier at the previous round. Our idea is that although the correspondence $\mathbf{c}_i$ is an outlier at the previous round, the less the residual of the correspondence at the current round is, the higher the true inlier probability of the potential global inlier $\mathbf{c}_i$ being included in $\mathbf{C}^{sul}$ is. Using the standard library in C++ \citep{schaling2011boost}, the true inlier probability of $\mathbf{c}_i$ is calculated using the following special case of Gamma distribution:
\begin{align} \label{eq:probability}
P(\mathbf{c}_i) &= 1 - \frac{\gamma\left(\frac{3}{2}, \frac{{r^{(R)}(\mathbf{c}_i)}^2}{2\sigma^2}\right)}{\Gamma\left(\frac{3}{2}\right)}
\end{align}
with
\begin{align}
\gamma(s, x) &= \int_0^x t^{s-1} e^{-t} \, dt\\
\Gamma(s) &= \int_0^\infty t^{s-1} e^{-t} \, dt
\end{align}

\noindent where ``3'' denotes the space dimension in which the residual is calculated. $\gamma$ and $\Gamma$ refer to the lower incomplete gamma function and gamma function, respectively. Here, $\sigma$ is uniformly distributed over the interval [0, $T_r$], where $T_r$ is the specified residual threshold.

Figure~\ref{fig:gamma_fig} illustrates the true inlier probability of the potential global inlier $\mathbf{c}_i$ in which the x-axis denotes the residual of $\mathbf{c}_i$ at the current round. From Figure~\ref{fig:gamma_fig}, it indicates that the higher the value of $P(\mathbf{c}_i)$ is, the less the value of the residual $r^{(R)}(\mathbf{c}_i)$ is. On the other hand, the higher the value of $P(\mathbf{c}_i)$, the higher the probability that the potential global correspondence $\mathbf{c}_i$ is truly included in $\mathbf{C}^{sul}$.

Using the calculated true inlier probability of $\mathbf{c}_i$ by Equation~(\ref{eq:probability}), denoted by $P(\mathbf{c}_i)$, we want to propose the second rule to assure that a potential global inlier $\mathbf{c}_i$ can be included in the local SUL correspondence set $\mathbf{C}^{sul}$. Before that, we introduce the definition of the randomly selected Mersenne's probability threshold~\citep{1998Mersenne}, which will be used as a dynamic threshold, not static threshold, to help determining whether $\mathbf{c}_i$ can be truly included in $\mathbf{C}^{sul}$ or not. The purpose of adopting Mersenne’s probability threshold is twofold: (1) our idea described below Rule~\ref{rule:R1} is kept, and (2) the randomness sense is employed into the inclusion decision: whether a potential global inlier $\mathbf{c}_i$ can be included in $\mathbf{C}^{sul}$.

\begin{definition}\label{def:D2}\sl{Given an interval, say [1, 100], we first select a number $n$ from the interval randomly, then the randomly selected Mersenne's probability threshold equals $p = \frac{n}{100}$.}
\end{definition}

Suppose the value of $P(\mathbf{c}_i)$ is 0.1. By Definition~\ref{def:D2}, if the randomly selected number is $n = 20$, the Mersenne's probability threshold is 0.2. Due to $P(\mathbf{c}_i)\,(=0.1) < 0.2$, the potential global inlier $\mathbf{c}_i$ cannot be included in $\mathbf{C}^{sul}$. From sampling viewpoint, due to $P(\mathbf{c}_i) = 0.1$, there is only a 10\% chance of selecting this number $n$ such that the condition ``$p\ (= \frac{n}{100}) \leq 0.1$'' holds. Therefore, it makes sense to utilize Mersenne’s probability threshold to increase the randomness when determining whether a potential global correspondence $\mathbf{c}_i$ can be truly included in $\mathbf{C}^{sul}$ or not.

\begin{myrule}\label{rule:R2}\sl{For a correspondence $\mathbf{c}_i$, it is a potential global inlier at the $R$-th interaction round, but is an outlier at the $(R - 1)$-th round. Based on the calculated true inlier probability of $\mathbf{c}_i$ by Equation~(\ref{eq:probability}), denoted by $P(\mathbf{c}_i)$, the correspondence $\mathbf{c}_i$ can be included in the local SUL correspondence set $\mathbf{C}^{sul}$ if the value of $P(\mathbf{c}_i)$ is greater than the randomly selected Mersenne's probability threshold.}
\end{myrule}

Rule~\ref{rule:R2} not only can preserve the principle: the higher the value of $P(\mathbf{c}_i)$, the higher the probability that the potential global correspondence $\mathbf{c}_i$ is truly included in $\mathbf{C}^{sul}$, but it is also with randomness.

\subsubsection{Determine whether one unreliable local correspondence $\mathbf{c}_i$ in $\mathbf{C}^{sul}$ should be removed or not}

After GLO-RANSAC updates $\mathbf{T}^{glo}$, ${Ir}^{glo}$, $t^{glo}$, and $\mathbf{Cl}^{glo}$ at the end of each RANSAC interaction round, if the condition ``$\mathbf{Cl}^{glo} < 99.5\%$ and $R < 5$'' holds, we also want to determine whether one unreliable local correspondence $\mathbf{c}_i$ in $\mathbf{C}^{sul}$ must be removed.

\begin{definition}\label{def:D3}\sl{One local correspondence $\mathbf{c}_i$ is in $\mathbf{C}^{sul} \backslash{Ir}^{glo}$, where ${Ir}^{lcl}$ is the local inlier set, $\mathbf{c}_i$ is called a potential local outlier.}
\end{definition}

By Definition~\ref{def:D3}, for each potential local outlier $\mathbf{c}_i$, if the condition ``$r^{(R-1)}(\mathbf{c}_i) \geq T_r$ and $r^{(R)}(\mathbf{c}_i) \geq T_r$'' holds, it means that $\mathbf{c}_i$ is always an outlier at the current and previous rounds. We thus treat $\mathbf{c}_i$ as a true outlier and remove $\mathbf{c}_i$ from $\mathbf{C}^{sul}$. Otherwise, if the condition ``$r^{(R-1)}(\mathbf{c}_i) < T_r$ and $r^{(R)}(\mathbf{c}_i) \geq T_r$'' holds, before treating $\mathbf{c}_i$ as a true outlier to be removed from $\mathbf{C}^{sul}$, the true outlier probability of $\mathbf{c}_i$ is calculated using Equation~(\ref{eq:probability}). Let the calculated true outlier probability of $\mathbf{c}_i$ be denoted by $P(\mathbf{c}_i)$. The following rule is used to determine whether a potential local outlier $\mathbf{c}_i$ should be removed or not.

\begin{myrule}\label{rule:R3}\sl{For a potential local outlier $\mathbf{c}_i$, if the value of $(1-P(\mathbf{c}_i))$ is greater than the randomly selected Mersenne’s probability $p$, $\mathbf{c}_i$ is truly removed from $\mathbf{C}^{sul}$.}
\end{myrule}

\noindent Rule~\ref{rule:R3} indicates that the higher the value of $(1-P(\mathbf{c}_i))$ is, the lower the value of $P(\mathbf{c}_i)$ is, implying that the higher the residual of $\mathbf{c}_i$, the higher the probability of $\mathbf{c}_i$ being removed from $\mathbf{C}^{sul}$.

\subsection{The Pseudo Code of Our Whole PCR Algorithm}
After describing all components of the proposed PCR algorithm, the pseudo code of our whole algorithm, called Algorithm~\ref{alg:ours}, is listed below.

\begin{breakablealgorithm}
\caption{Our registration algorithm}
\begin{algorithmic}[1]
\label{alg:ours}

\textbf{Input:} Initial correspondence and line vector sets, $\mathbf{C}$ and $\mathbf{L}$

\textbf{Output:} Registration solution $\mathbf{T}^{glo}$

\STATE Construct $\mathbf{C}^{sul}$ from $\mathbf{C}$ via the AHS-based method
\STATE Construct $\mathbf{L}^{sul}$ from $\mathbf{C}^{sul}$ via the LVLP-based method
\STATE $\mathbf{W} \leftarrow \mathbf{0}$; $R \leftarrow 0$; $t^{glo} \leftarrow 0$; $\mathbf{T}^{glo} \leftarrow I_{4 \times 4}$
\WHILE{\textbf{true}}
    \STATE Sample $\mathbf{L}^{sul}_{sub}$ from $\mathbf{L}^{sul}$
    \STATE $t^{lcl} \leftarrow 0$
    \REPEAT
        \STATE Sample $\mathbf{L}^{sul}_{basic}$ from $\mathbf{L}^{sul}_{sub}$
        \STATE Estimate $\mathbf{T}^{lcl}$ via TEASER++ using the received $\mathbf{T}^{glo}$ as the initial local transformation
        \STATE $t^{lcl} \leftarrow t^{lcl} + 1$
        \IF{Equation~(\ref{eq:inherit}) holds and $\mathbf{T}^{lcl} \approx \mathbf{T}^{glo}$}
            \STATE $t^{lcl} \leftarrow t^{glo} + t^{lcl}$
            \STATE \textbf{break}
        \ENDIF
        \STATE $\mathbf{Cl}^{lcl} \leftarrow 1 - (1 - |Ir^{lcl}|)^{t^{lcl}}$
    \UNTIL{$\mathbf{Cl}^{lcl} \geq 99.5\%$}
    \STATE Update $\mathbf{T}^{glo}$, $t^{glo}$, $Ir^{glo}$, and $\mathbf{Cl}^{glo}$ using the received $\mathbf{T}^{lcl}$ and $t^{lcl}$
    \IF{$\mathbf{Cl}^{glo} \geq 99.5\%$ or $R = 5$}
        \STATE \textbf{break}
    \ELSE
        \STATE Update $\mathbf{W}$ by incrementing the weights of correspondences in $Ir^{glo}$
        \STATE Update $\mathbf{C}^{sul}$ and $\mathbf{L}^{sul}$ via the proposed probabilistic SUS
        \STATE $R \leftarrow R + 1$
    \ENDIF
\ENDWHILE
\STATE Compute the final transformation $\mathbf{T}^{glo}$ via the weighted SVD on $\mathbf{C}$ associated with $\mathbf{W}$
\STATE \textbf{return} $\mathbf{T}^{glo}$

\end{algorithmic}
\end{breakablealgorithm}

After presenting the pseudo code of our whole PCR algorithm, in the next section, the comprehensive experimental data are demonstrated to justify the registration accuracy and time merits of our algorithm relative to the state-of-the-art methods.

\section{Experimental Results}\label{sec:experiment}
In this section, we demonstrate the registration accuracy and time merits of our algorithm relative to the eight state-of-the-art methods: GC-RANSAC \citep{barath2021gcransac}, RANSIC \citep{sun2021ransic}, One-point RANSAC \citep{li2021onepoint}, TEASER++ \citep{yang2020teaser}, $\text{SC}^2$-PCR++~\citep{chen2023sc2pcr}, MAC++~\citep{zhang2025mac++}, MAGSC~\citep{xu2025globally}, and TEAR \citep{Huang2024tear}. The accuracy of each considered method is measured by seven objective quality metrics, which are defined in Equations~(\ref{eq:R_error})-(\ref{eq:P_R_F1}). The subjective quality comparison of some representative methods is also provided. In addition, in terms of our probabilistic SUS and the global early termination condition, the ablation study of our algorithm relative to the baseline method \citep{wt2024cransac} is provided.

The runtime of each considered method is measured in seconds. Note that for time comparison fairness, the time required in constructing the line vector set $\mathbf{L}$ in TEASER++, One-point RANSAC, RANSIC, $\text{SC}^2$-PCR++, and our algorithm is included in the total runtime. To evaluate the registration accuracy of each considered method, seven accuracy metrics, namely the rotation error ($\mathbf{R}^{err}$), translation error ($\mathbf{T}^{err}$), RMSE, MeSE, precision (P), recall (R), and F1-score (F1), are used.

Let ($\mathbf{R}^{gt}$, $\mathbf{T}^{gt}$) denote the ground truth transformation; let ($\bm{R}^{est}$, $\bm{T}^{est}$) denote the estimated transformation obtained by one considered method. The first four accuracy metrics are defined by
\begin{equation} \label{eq:R_error}
{\mathbf{R}^{err}} = \arccos\left(\frac{tr({R}^{gt}({R}^{est})^T)-1}{2}\right)
\end{equation} 
\begin{equation} \label{eq:t_error}
{\mathbf{T}^{err}} = \|T^{gt} - T^{est}\|
\end{equation} 
\begin{equation} \label{eq:RMSE_error} 
\text{RMSE} = \sqrt{\frac{1}{|{\mathbf{P}^s}|}\sum_{i=1}^{|{\mathbf{P}^s}|}\lVert x^{gt}_i - x^{est}_i \rVert^2}
\end{equation}
\begin{equation} \label{eq:MeSE_error}
\text{MeSE} = \text{Median}\left(
\begin{aligned}
&\|x_1^{gt} - x_1^{est}\|, \|x_2^{gt} - x_2^{est}\|, \ldots, \\
&\|x_{|\mathbf{P}^s|}^{gt} - x_{|\mathbf{P}^s|}^{est}\|
\end{aligned} \right)
\end{equation}
where $x^{gt}_i$ and $x^{est}_i$ represent the transformed source points by performing the ground truth transformation and the estimated transformation on the source point $x_i \in \mathbf{P}^s$, respectively.

Let the number of correctly recognized inliers be denoted by TP; the number of outliers incorrectly recognized as inliers be denoted by FP; the number of inliers incorrectly recognized as outliers be denoted by FN. The larger the value of precision is, the less the number of outliers incorrectly classified as inliers is. The larger the value of recall is, the less the number of inliers incorrectly classified as outliers is. F1-score is a compromise of precision and recall. The last three accuracy metrics are defined by
\begin{equation}
\begin{aligned}
\text{Precision} &= \frac{\text{TP}}{\text{TP}+\text{FP}} \\
\text{Recall} &= \frac{\text{TP}}{\text{TP}+\text{FN}} \\
\text{F1-score} &= 2 \times \frac{\text{Precision} \times \text{Recall}}{\text{Precision} + \text{Recall}}
\end{aligned}
\label{eq:P_R_F1}
\end{equation}

The codes of all comparative methods are available. One-point RANSAC and RANSIC are implemented in Matlab. The other six comparative methods are implemented in C++. The C++ source code of our algorithm is available at \href{https://github.com/ivpml84079/Probabilistic-Self-Update-Line-Vector-Set-Based-Point-Cloud-Registration}{https://github.com/ivpml84079/Probabilistic-Self-Update-Line-Vector-Set-Based-Point-Cloud-Registration}. To cover various scenarios, all considered methods are performed on the three public datasets: 3DMatch, KITTI, and WHU-TLS. For fairness, all considered methods are implemented on the same machine equipped with a computer with an AMD Ryzen 7 3800X Processor, 40GB RAM, and the GeForce RTX 3080 Ti GPU.

\begin{figure*}[htbp] 
\captionsetup{labelsep=period}
\centering

\begin{minipage}[t]{0.48\textwidth}
    \centering
    \includegraphics[width=\textwidth]{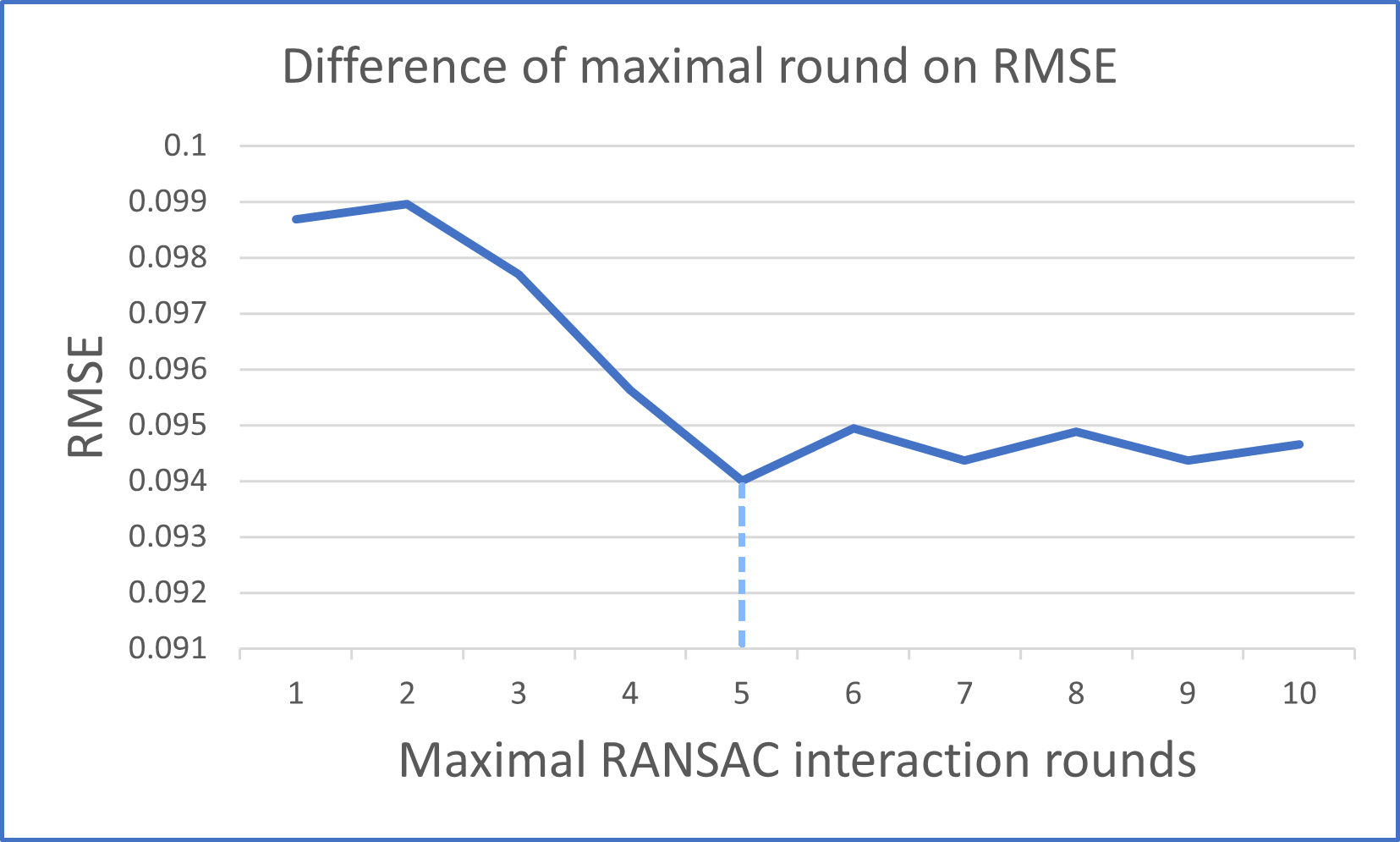}
    \caption*{(a)}
    \label{fig:QR_RMSE}
\end{minipage}\hfill
\begin{minipage}[t]{0.48\textwidth}
    \centering
    \includegraphics[width=\textwidth]{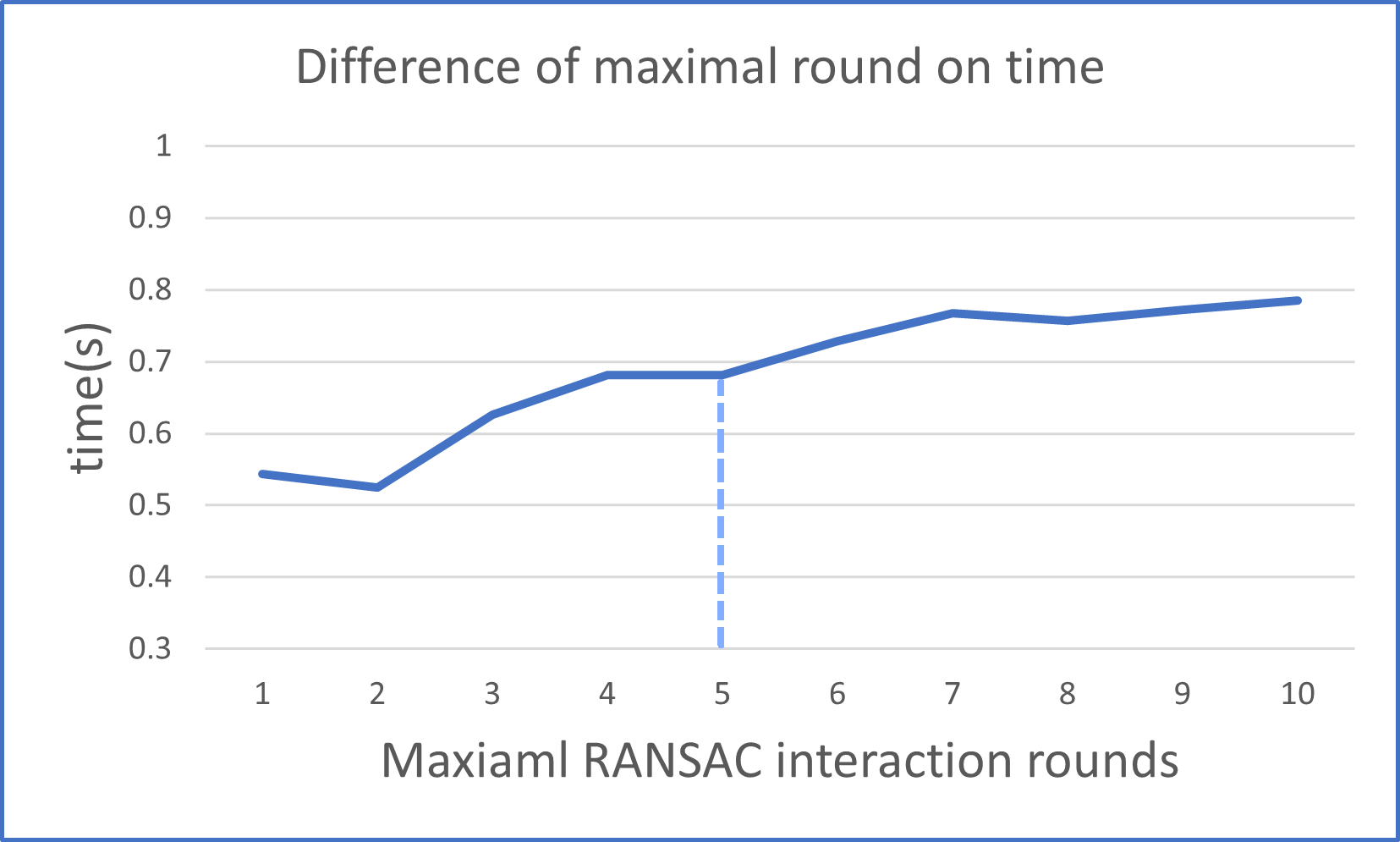}
    \caption*{(b)}
    \label{fig:QR_time}
\end{minipage}

\caption{The choice of the maximal RANSAC interaction rounds, i.e. $R_{max}$, as our global early termination condition. (a) The RMSE plot. (b) The time plot.}
\label{fig:QR}
\end{figure*} 

\subsection{Testing Datasets Description and Parameters Setting}
\subsubsection{Testing dataset description}
In the 3DMatch dataset which mainly contains indoor scenes, there are 1623 point cloud pairs and they are available at \href{https://3dmatch.cs.princeton.edu/}{https://3dmatch.cs.princeton.edu}. In the KITTI dataset which mainly contains outdoor scenes, there are 555 point cloud pairs and they are available at \href{https://www.cvlibs.net/datasets/kitti/}{https://www.cvlibs.net/datasets/kitti}. In 3DMatch and KITTI, each point cloud has been down-sampled to about 5000 points. The WHU-TLS dataset is available at \href{https://3s.whu.edu.cn/ybs/en/benchmark.htm}{https://3s.whu.edu.cn/ybs/en/benchmark.htm}. In our experiments, we use the WHU-TLS subset with two scenes (Park and Mountain), totaling 36 point-cloud pairs.

\subsubsection{Parameter Setting} \label{sec:parameter}
Following most of the existing methods, the residual threshold $T_r$ is set to 0.01 for the 3DMatch dataset; 0.1 for the KITTI dataset; 0.2 for the WHU-TLS dataset.

As to determine the choice of the number maximal allowable of RANSAC interaction rounds, i.e., $R_{max}$, as our global early termination condition, according to the three testing datasets, we have tried the parameter interval: $1 \leq R_{max} \leq 10$. Figure~\ref{fig:QR}(a) and Figure~\ref{fig:QR}(b) depict the RMSE plot and the time plot, respectively, using our PCR algorithm. From Figure~\ref{fig:QR}, we observe that when the RANSAC interaction round parameter is larger than five, the RMSE value remains the same small value, but it takes more time; when the interaction round parameter is slightly smaller than five, it causes significantly larger RMSE, but it takes slightly less time. Therefore, $R_{max}$ is set to five.

\begin{figure*}[htbp] 
\captionsetup{labelsep=period}
\centering

\begin{minipage}[t]{0.3\textwidth}
    \centering
    \includegraphics[width=\textwidth]{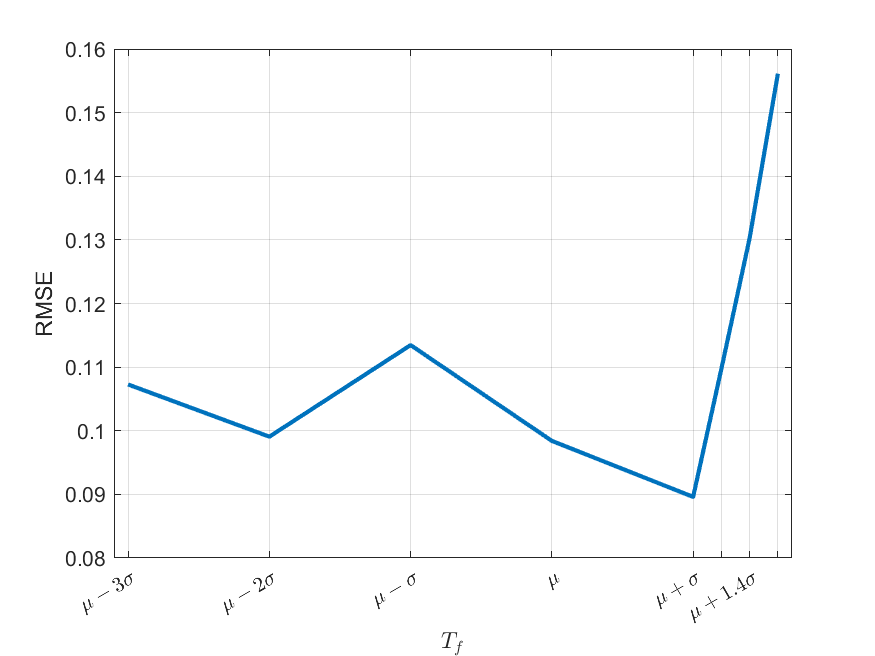}
    \caption*{(a)}
\end{minipage}\hfill
\begin{minipage}[t]{0.3\textwidth}
    \centering
    \includegraphics[width=\textwidth]{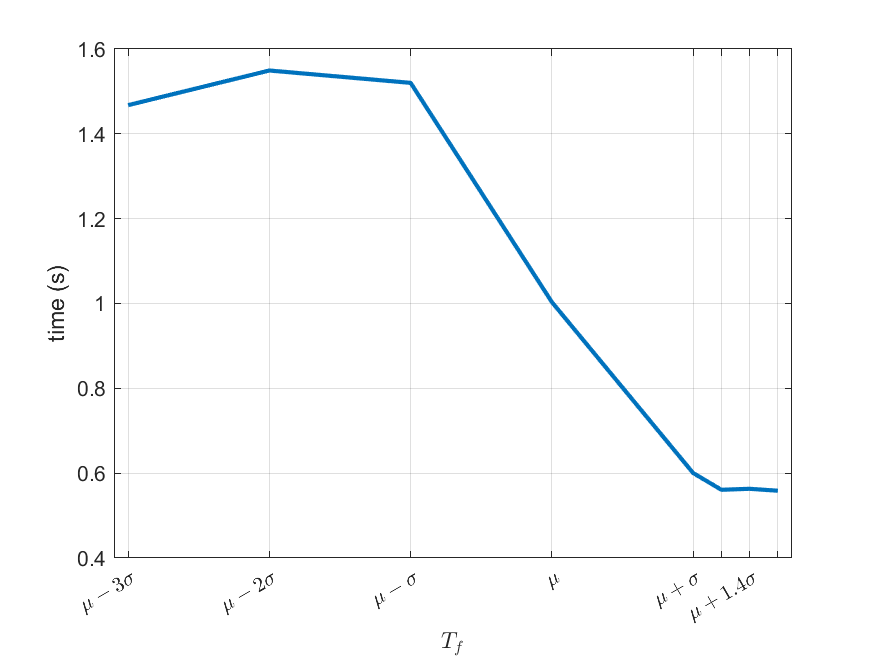}
    \caption*{(b)}
\end{minipage}\hfill
\begin{minipage}[t]{0.3\textwidth}
    \centering
    \includegraphics[width=\textwidth]{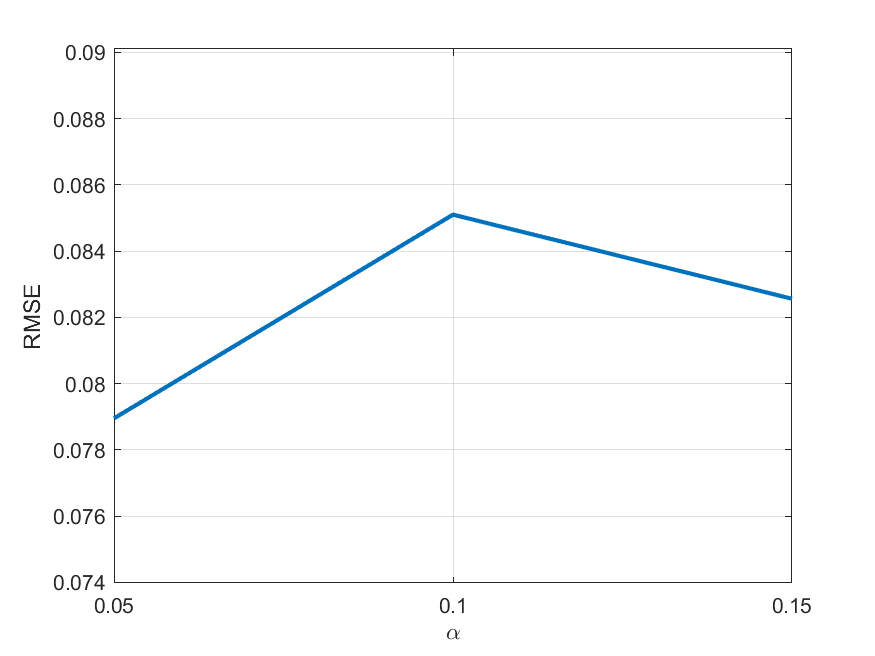}
    \caption*{(c)}
\end{minipage}

\vspace{10pt}

\begin{minipage}[t]{0.3\textwidth}
    \centering
    \includegraphics[width=\textwidth]{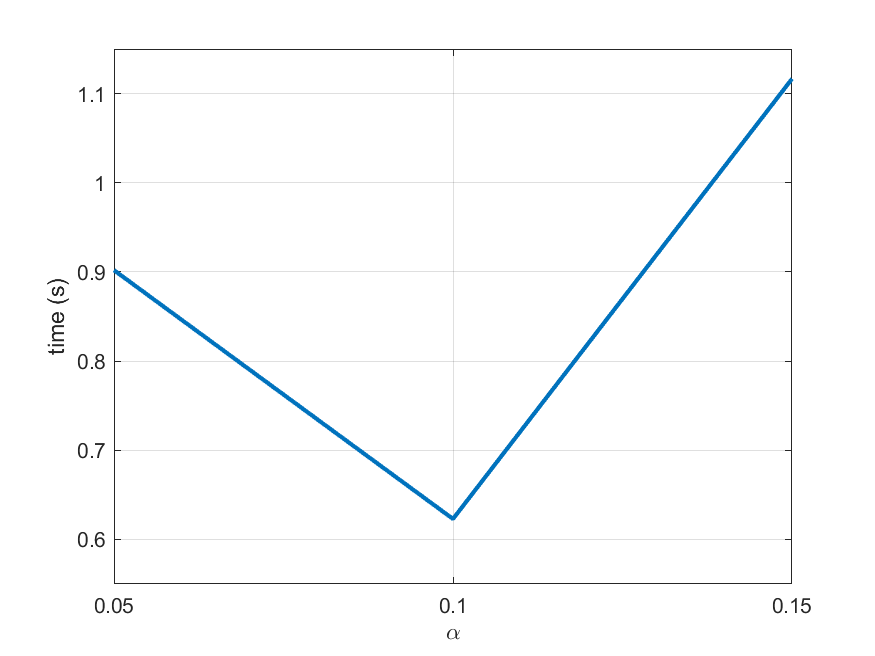}
    \caption*{(d)}
\end{minipage}\hfill
\begin{minipage}[t]{0.3\textwidth}
    \centering
    \includegraphics[width=\textwidth]{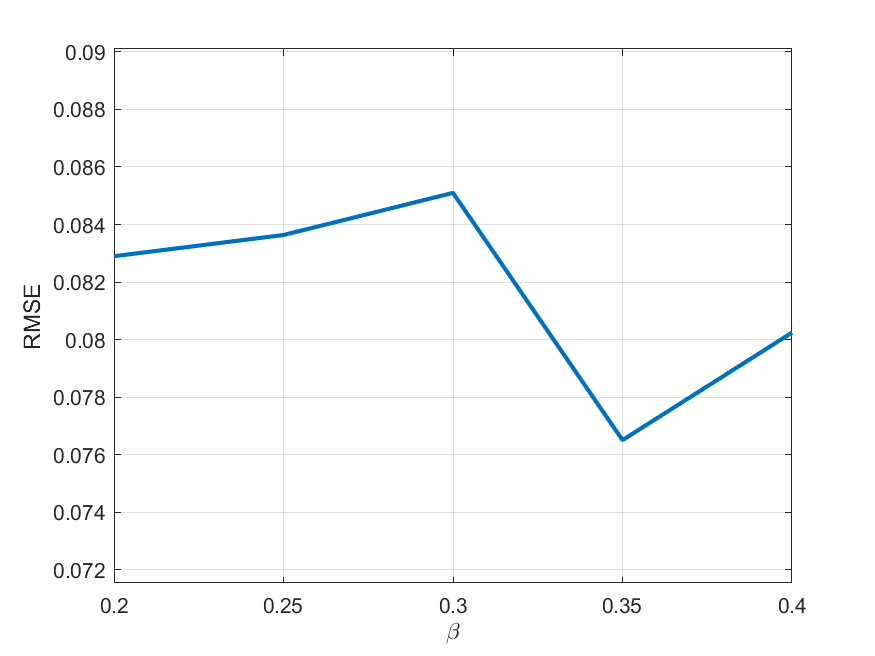}
    \caption*{(e)}
\end{minipage}\hfill
\begin{minipage}[t]{0.3\textwidth}
    \centering
    \includegraphics[width=\textwidth]{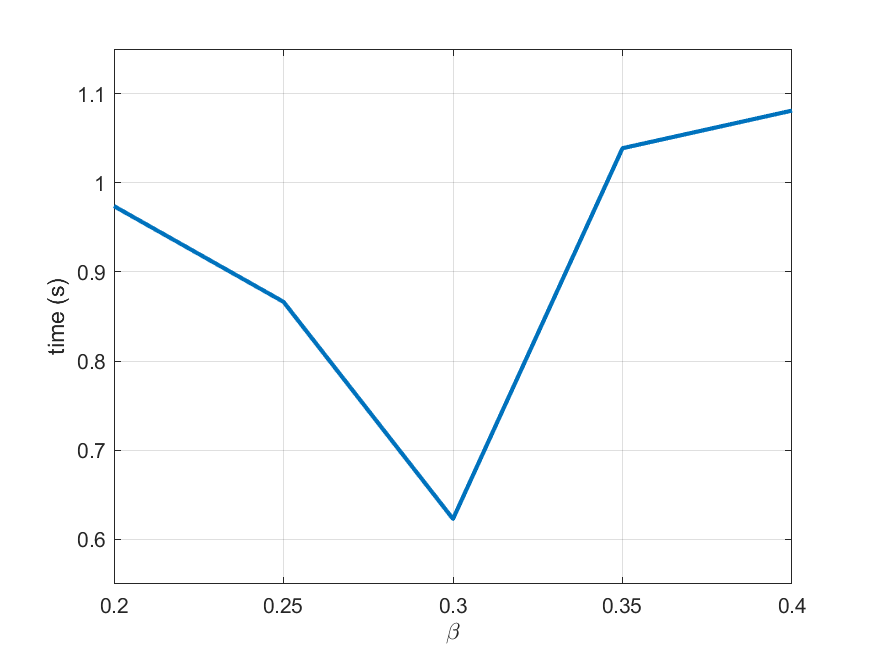}
    \caption*{(f)}
\end{minipage}

\caption{Determine the specified values of $T_f$, $\alpha$, and $\beta$. (a) RMSE diagram with respect to $T_f$. (b) Execution time diagram with respect to $T_f$. (c) RMSE diagram with respect to $\alpha$. (d) Execution time diagram with respect to $\alpha$. (e) RMSE diagram with respect to $\beta$. (f) Execution time diagram with respect to $\beta$.}
\label{fig:setting}
\end{figure*} 

To select the value of the frequency threshold $T_f$ for constructing the initial SUL correspondence set $\mathbf{C}^{sul}$, we have tried the parameter interval: $\mu-3\sigma \le T_f \le \mu+1.6\sigma$. As shown in Figures~\ref{fig:setting}(a)-(b), $T_f$ = $\mu+\sigma$ is the best choice for achieving the best balance between RMSE and time. In the two random subset selection processes of LCL-RANSAC, after trying the parameter intervals: $5 \leq \alpha \leq 15$ and $20 \leq \beta \leq 40$, setting $\alpha = 10$ and $\beta = 30$ is a good choice for achieving better accuracy and time.

\subsection{Registration Accuracy and Execution Time Comparison}
We first compare the registration accuracy and time among all considered PCR methods. Then, the visual quality of our algorithm is illustrated. Additionally, in terms of our probabilistic SUS and the global early termination condition, the ablation study of our algorithm is provided.

\begin{table*}[p] 
    \centering
    \begin{minipage}{15.6cm} 
        \captionsetup{justification=raggedright, singlelinecheck=false, labelsep=newline}
        \caption{The registration accuracy and execution time comparison.}
        \label{tab:acc_time}
        
        \footnotesize 
        \renewcommand{\arraystretch}{0.85} 
        \setlength{\tabcolsep}{6pt} 
        
        \begin{tabular}{
            >{\raggedright\arraybackslash}p{3cm} 
            >{\centering\arraybackslash}p{2cm}   
            >{\centering\arraybackslash}p{2cm}   
            >{\centering\arraybackslash}p{2cm}   
            >{\centering\arraybackslash}p{2cm}   
            >{\centering\arraybackslash}p{2cm}   
        } \toprule
             Method & Metric & 3DMATCH & KITTI & WHU-TLS & Average \\ \midrule
             \multirow{8}{*}{GC-RANSAC \citep{barath2021gcransac}}
             & ${\rm{R}}^{err}$ (\textdegree) & 2.211 & 0.941 & 0.123 & 1.092 \\
             & $T^{err}$ (cm) & 7.090 & 15.755 & \textbf{4.671} & 9.172 \\
             & RMSE (m) & 0.050 & 0.345 & 0.168 & 0.188 \\
             & MeSE (m) & \tcblu{0.046} & 0.246 & 0.126 & 0.139 \\
             & P (\%) & 85.457 & 90.054 & \tcorg{99.056} & 91.522 \\
             & R (\%) & 91.136 & 82.601 & 95.222 & 89.653 \\
             & F (\%) & 88.015 & 85.817 & 96.614 & 90.149 \\
             & Time (s) & 0.576 & 0.493 & 1.895 & 0.988 \\ \midrule
             
             \multirow{8}{*}{TEASER++ \citep{yang2020teaser}}
             & ${\rm{R}}^{err}$ (\textdegree) & 2.421 & 0.655 & 0.187 & 1.088 \\
             & $T^{err}$ (cm) & 7.986 & 11.861 & 8.870 & 9.572 \\
             & RMSE (m) & 0.055 & 0.220 & 0.234 & 0.170 \\
             & MeSE (m) & 0.051 & 0.163 & 0.168 & 0.127 \\
             & P (\%) & \tcorg{85.966} & 92.317 & 97.767 & 92.017 \\
             & R (\%) & 88.020 & 90.026 & 97.754 & 91.933 \\
             & F1 (\%) & 86.780 & 91.061 & 97.604 & 91.815 \\
             & Time (s) & 1.789 & 0.710 & 0.812 & 1.104 \\ \midrule
             
             \multirow{8}{*}{\shortstack[l]{One-point\\RANSAC~\citep{li2021onepoint}}} 
             & ${\rm{R}}^{err}$ (\textdegree) & 2.833 & 0.474 & 0.121 & 1.143 \\
             & $T^{err}$ (cm) & 8.499 & \tcblu{9.208} & 6.279 & 7.995 \\
             & RMSE (m) & 0.061 & 0.180 & 0.149 & 0.130 \\
             & MeSE (m) & 0.055 & 0.135 & \tcblu{0.110} & 0.100 \\
             & P (\%) & 84.133 & 91.831 & 96.920 & 90.961 \\
             & R (\%) & 87.717 & 93.728 & 98.650 & 93.365 \\
             & F1 (\%) & 85.599 & 92.705 & 97.738 & 92.014 \\
             & Time (s) & 3.136 & 2.502 & 1.661 & 2.433 \\ \midrule
             
             \multirow{8}{*}{RANSIC {\citep{sun2021ransic}}}
             & ${\rm{R}}^{err}$ (\textdegree) & 2.624 & 0.818 & 0.266 & 1.236 \\
             & $T^{err}$ (cm) & 7.807 & 16.517 & 12.928 & 12.417 \\
             & RMSE (m) & 0.057 & 0.326 & 0.343 & 0.242 \\
             & MeSE (m) & 0.052 & 0.242 & 0.257 & 0.184 \\
             & P (\%) & 83.434 & 88.242 & 95.870 & 89.182 \\
             & R (\%) & 89.336 & 83.679 & 91.714 & 88.243 \\
             & F1 (\%) & 86.090 & 85.674 & 93.116 & 88.293 \\
             & Time (s) & \textbf{0.023} & \textbf{0.079} & 3.379 & 1.160 \\ \midrule
             
             \multirow{8}{*}{$\text{SC}^2$-PCR++ {\citep{chen2023sc2pcr}}}
             & ${\rm{R}}^{err}$ (\textdegree) & \tcorg{2.055} & \tcblu{0.416} & 0.135 & 0.869 \\
             & $T^{err}$ (cm) & \tcorg{6.593} & 9.778 & 8.201 & 8.191 \\
             & RMSE (m) & \tcorg{0.048} & 0.162 & 0.182 & 0.131 \\
             & MeSE (m) & \textbf{0.044} & \tcblu{0.127} & 0.129 & 0.100 \\
             & P (\%) & 84.607 & 91.802 & 97.096 & 91.168 \\
             & R (\%) & 91.712 & \tcblu{94.259} & 98.585 & \tcblu{94.852} \\
             & F1 (\%) & 87.813 & 92.953 & 97.798 & 92.855 \\
             & Time (s) & \tcblu{0.389} & \tcorg{0.083} & \tcorg{0.114} & \tcorg{0.195} \\ \midrule
             
             \multirow{8}{*}{TEAR {\citep{Huang2024tear}}}
             & ${\rm{R}}^{err}$ (\textdegree) & \textbf{1.764} & 0.445 & \tcorg{0.090} & \textbf{0.766} \\
             & $T^{err}$ (cm) & \textbf{6.169} & 10.666 & \tcblu{5.184} & \tcblu{7.340} \\
             & RMSE (m) & \textbf{0.045} & \tcblu{0.158} & 0.154 & \tcblu{0.119} \\
             & MeSE (m) & \textbf{0.044} & 0.158 & 0.152 & 0.118 \\
             & P (\%) & \tcblu{85.681} & 91.634 & \textbf{99.313} & \tcorg{92.209} \\
             & R (\%) & \tcorg{92.145} & 93.254 & \textbf{98.735} & 94.711 \\
             & F1 (\%) & \tcorg{88.613} & 92.343 & \textbf{99.022} & \tcblu{93.326} \\
             & Time (s) & 0.640 & 0.907 & 0.830 & \tcblu{0.792} \\ \midrule
             
             \multirow{8}{*}{MAC++{\citep{zhang2025mac++}}}
             & ${\rm{R}}^{err}$ (\textdegree) & \tcblu{2.072} & \tcorg{0.362} & \tcorg{0.090} & \tcorg{0.811} \\
             & $T^{err}$ (cm) & \tcblu{6.687} & \tcorg{8.135} & 7.190 & \tcorg{7.337} \\
             & RMSE (m) & \tcorg{0.048} & \tcorg{0.139} & \tcorg{0.139} & \tcorg{0.108} \\
             & MeSE (m) & \textbf{0.044} & \tcorg{0.108} & \tcblu{0.110} & \tcorg{0.087} \\
             & P (\%) & 85.431 & \tcorg{93.209} & 97.938 & \tcblu{92.193} \\
             & R (\%) & 91.958 & \textbf{95.082} & \tcblu{98.675} & \tcorg{95.238} \\
             & F1 (\%) & \tcblu{88.374} & \tcorg{94.076} & 98.267 & \tcorg{93.572} \\
             & Time (s) & 3.808 & 2.185 & 1.175 & 2.389 \\ \midrule
             
             \multirow{8}{*}{MAGSC{\citep{xu2025globally}}}
             & ${\rm{R}}^{err}$ (\textdegree) & 2.139 & 0.456 & \tcblu{0.102} & 0.899 \\
             & $T^{err}$ (cm) & 6.725 & 9.447 & 6.685 & 7.619 \\
             & RMSE (m) & \tcblu{0.049} & 0.167 & \tcblu{0.140} & \tcblu{0.119} \\
             & MeSE (m) & \tcorg{0.045} & \tcblu{0.127} & \tcorg{0.109} & \tcblu{0.093} \\
             & P (\%) & 84.544 & \tcblu{92.886} & \tcblu{98.528} & 91.986 \\
             & R (\%) & \tcblu{92.008} & 93.224 & 98.502 & 94.578 \\
             & F1 (\%) & 87.898 & \tcblu{92.976} & \tcblu{98.461} & 93.112 \\
             & Time (s) & 3.782 & \tcblu{0.237} & \tcblu{0.510} & 1.510 \\ \midrule
             
             \multirow{8}{*}{Ours}
             & ${\rm{R}}^{err}$ (\textdegree) & 2.139 & \textbf{0.278} & \textbf{0.051} & \tcblu{0.823} \\
             & $T^{err}$ (cm) & 6.913 & \textbf{7.729} & \tcorg{5.083} & \textbf{6.575} \\
             & RMSE (m) & \tcblu{0.049} & \textbf{0.117} & \textbf{0.083} & \textbf{0.083} \\
             & MeSE (m) & \textbf{0.044} & \textbf{0.096} & \textbf{0.066} & \textbf{0.069} \\
             & P (\%) & \textbf{91.719} & \textbf{94.973} & 98.523 & \textbf{95.072} \\
             & R (\%) & \textbf{94.013} & \tcorg{94.282} & \tcorg{98.728} & \textbf{95.674} \\
             & F1 (\%) & \textbf{92.736} & \textbf{94.567} & \tcorg{98.611} & \textbf{95.305} \\
             & Time (s) & \tcorg{0.176} & \tcorg{0.083} & \textbf{0.105} & \textbf{0.121} \\ \bottomrule
        \end{tabular}
    \end{minipage}
\end{table*}

\subsubsection{Quantitative quality and time comparison}

For the three testing dataset, Table~\ref{tab:acc_time} shows the average quantitative quality and time of each considered method in which the best, second-best, and third-best results are marked in black bold, orange and blue, respectively. For simplicity, Precision, Recall, and F1-score are abbreviated as P, R, and F1, respectively.

\vspace{1ex}
\noindent \quad a) For the 3DMatch dataset:
For the 3DMatch dataset, Table~\ref{tab:acc_time} shows the quantitative quality and runtime of each considered method. From Table~\ref{tab:acc_time}, we observe that relative to all comparative methods, our algorithm, abbreviated as ``Ours'', achieves the best performance in MeSE, P, R, and F1; the second-best performance in runtime (Time); the third-best performance in RMSE. The TEAR method achieves the best performance in ${\rm{R}}^{err}$, $T^{err}$, RMSE, and MeSE; the second-best performance in R and F1; the third-best performance in P. The $\text{SC}^2$-PCR++ method achieves the best performance in MeSE; the second-best performance in ${\rm{R}}^{err}$, $T^{err}$, and RMSE; the third-best performance in Time. The MAC++ method achieves the best performance in MeSE and the second-best in RMSE, and third-best performance in ${\rm{R}}^{err}$, $T^{err}$, and F1. The TEASER++ method achieves the second-best performance in P. The MAGSC method achieves the second-best performance in MeSE; the third-best performance in RMSE and R. The GC-RANSAC method achieves the third-best performance in MeSE. The RANSIC method achieves the best performance in Time.

Overall, for the 3DMATCH dataset, in terms of the seven quality metrics and runtime, our algorithm attains the highest average performance, followed by the TEAR and MAC++ methods.

\vspace{1ex}
\noindent \quad b) For the KITTI dataset:
For the KITTI dataset, Table~\ref{tab:acc_time} indicates that our algorithm achieves the best performance in ${\rm{R}}^{err}$, $T^{err}$, RMSE, MeSE, P, and F1; the second-best performance in R and Time. The MAC++ method achieves the best performance in R; the second-best performance in ${\rm{R}}^{err}$, $T^{err}$, RMSE, MeSE, P, and F1. The $\text{SC}^2$-PCR++ method achieves the second-best performance in Time; the third-best performance in ${\rm{R}}^{err}$, MeSE, and R. The MAGSC method achieves the third-best performance in MeSE, P, F1, and Time. The One-point RANSAC method achieves the third-best performance in $T^{err}$. The TEAR method achieves the third-best performance in RMSE. The RANSIC method achieves the best performance in Time.

Overall, for the KITTI dataset, in terms of the seven quality metrics, our algorithm attains the highest overall accuracy, followed by the MAC++ method. Furthermore, for this dataset, our algorithm is much faster than MAC++, and the runtime improvement ratio is 96.20\%. Notably, despite the relatively sparse structure and higher noise levels in KITTI, which typically lead to less reliable normal vector estimation and a higher proportion of outliers in the initial SUL correspondence set, our proposed strategy efficiently manages these challenges. As a result, our algorithm avoids a significant increase in the computational cost for iterative refinement, achieving an even faster runtime on KITTI than on 3DMATCH and WHU-TLS. As for the absolute runtime performance, the RANSIC method achieves the best performance.

\vspace{1ex}
\noindent \quad c) For the WHU-TLS dataset:
For the WHU-TLS dataset, Table~\ref{tab:acc_time} indicates that the TEAR method achieves the best performance in P, R, and F1; the second-best performance in ${\rm{R}}^{err}$ and $T^{err}$. Our algorithm is competitive to TEAR, achieving the best performance in ${\rm{R}}^{err}$, RMSE, MeSE, and Time; the second-best performance in $T^{err}$, R, and F1. Notably, for this dataset, our algorithm is much faster than TEAR, and the runtime improvement ratio is 87.35\%. 

In summary, the WHU-TLS dataset is more challenging than 3DMATCH and KITTI due to its sparse structure and heavy outlier contamination, yet our algorithm still maintains competitive accuracy and superior efficiency among all compared methods.

\vspace{1ex}
\noindent \quad d) For the three used datasets
After demonstrating the quality and runtime merits of our algorithm for individual datasets, we further present the average quality and time merits of our algorithm for the three used datasets.

From the last column of Table~\ref{tab:acc_time}, it is observed that our algorithm achieves the best performance in $T^{err}$, RMSE, MeSE, P, R, F1, and Time; the third-best performance in ${\rm{R}}^{err}$. The MAC++ method achieves the second-best performance in $T^{err}$, RMSE, MeSE, R, and F1; the third-best performance in P. The TEAR method achieves the best performance in ${\rm{R}}^{err}$; the second-best performance in P; the third-best performance in $T^{err}$, RMSE, F1, and Time. The $\text{SC}^2$-PCR++ method achieves the second-best performance in Time; the third-best performance in R. The MAGSC method achieves the third-best performance in RMSE and MeSE.

\begin{table*}[htbp] 
\captionsetup{justification=raggedright, singlelinecheck=false, labelsep=newline}
\centering
\caption{The visual comparison for six point cloud pairs.} 
\label{tab:showcase}
\renewcommand{\arraystretch}{1.5}
\setlength{\tabcolsep}{3pt} 

\begin{tabular}{%
  >{\centering\arraybackslash}m{3.2cm}| 
  *{6}{>{\centering\arraybackslash}m{0.12\textwidth}}} \toprule 
\multicolumn{1}{c|}{Point Cloud Pairs} & Ground truth & One-point RANSAC & $\text{SC}^2$-PCR++ & TEAR & MAC++ & Ours \\ \midrule

\vspace{0.3cm}
3DMatch\_home\_md 
\vspace{0.2cm}& 
\vspace{0.1cm}\includegraphics[width=0.12\textwidth]{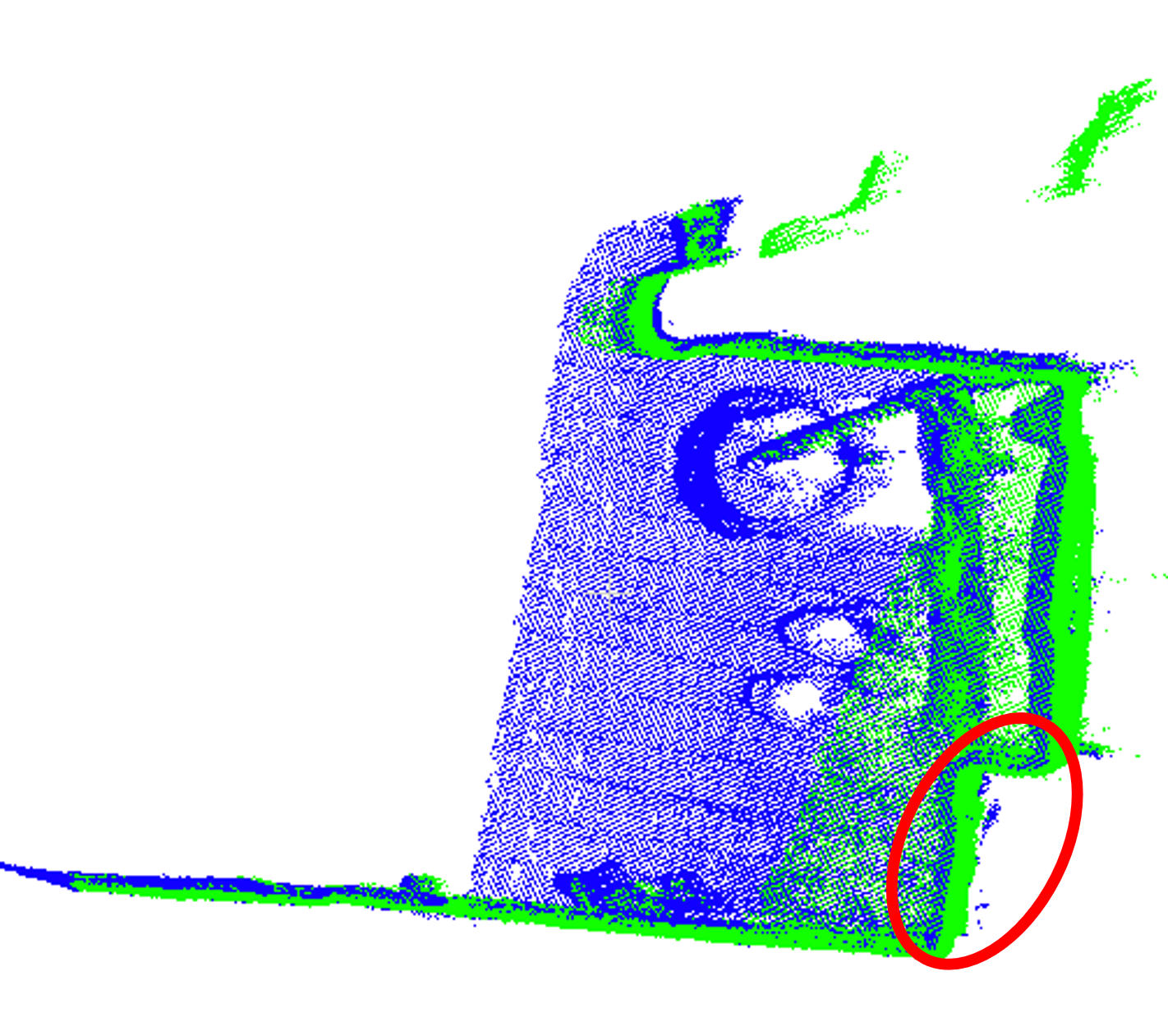} &
\vspace{0.1cm}\includegraphics[width=0.12\textwidth]{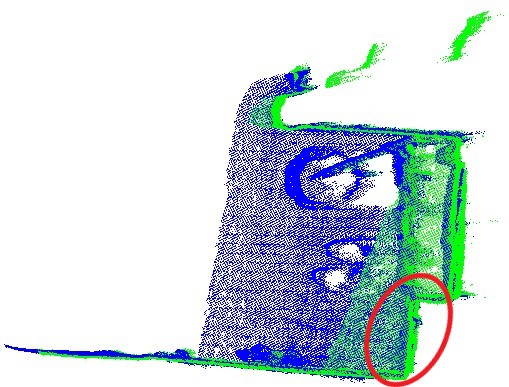} &
\vspace{0.1cm}\includegraphics[width=0.12\textwidth]{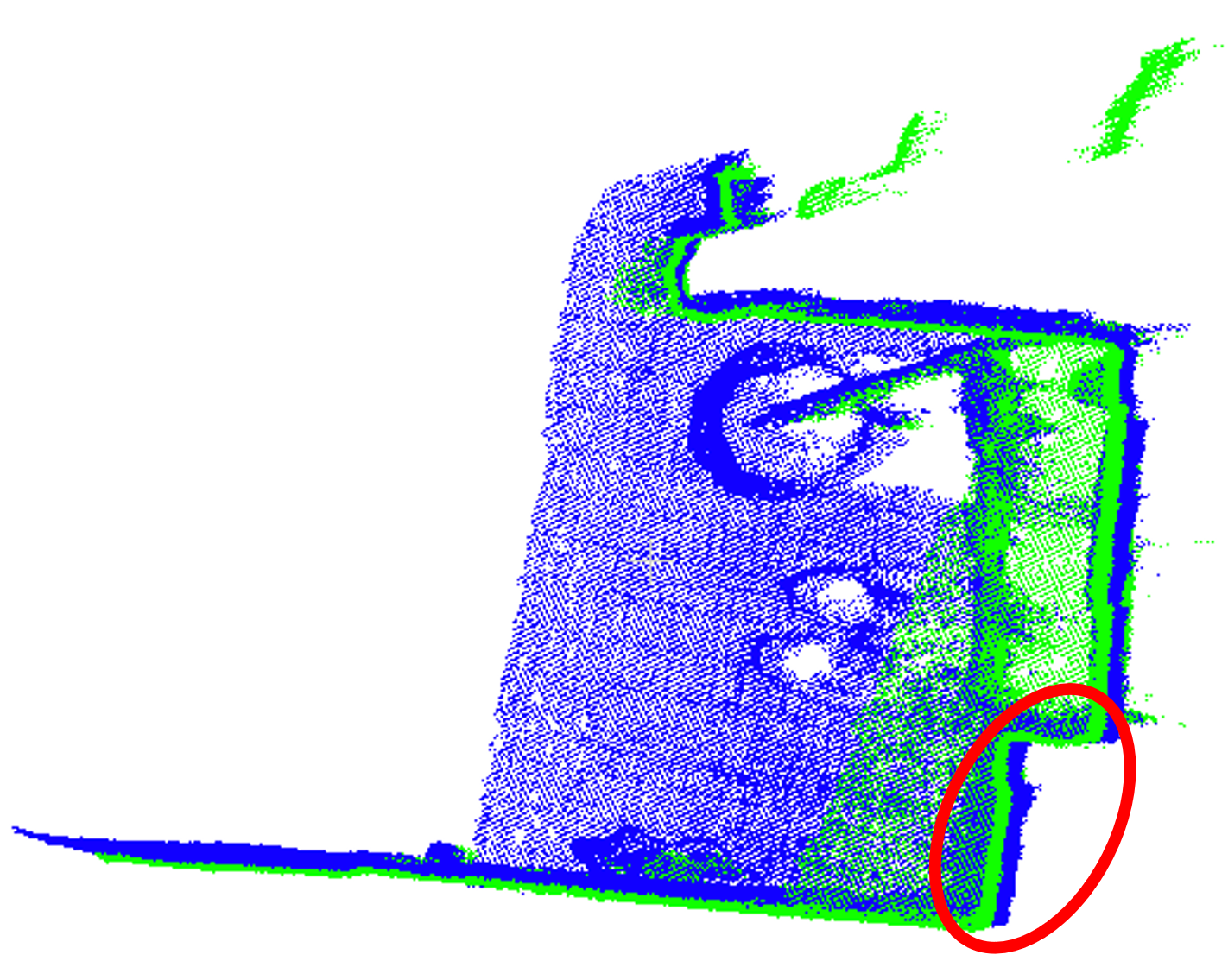} &
\vspace{0.1cm}\includegraphics[width=0.12\textwidth]{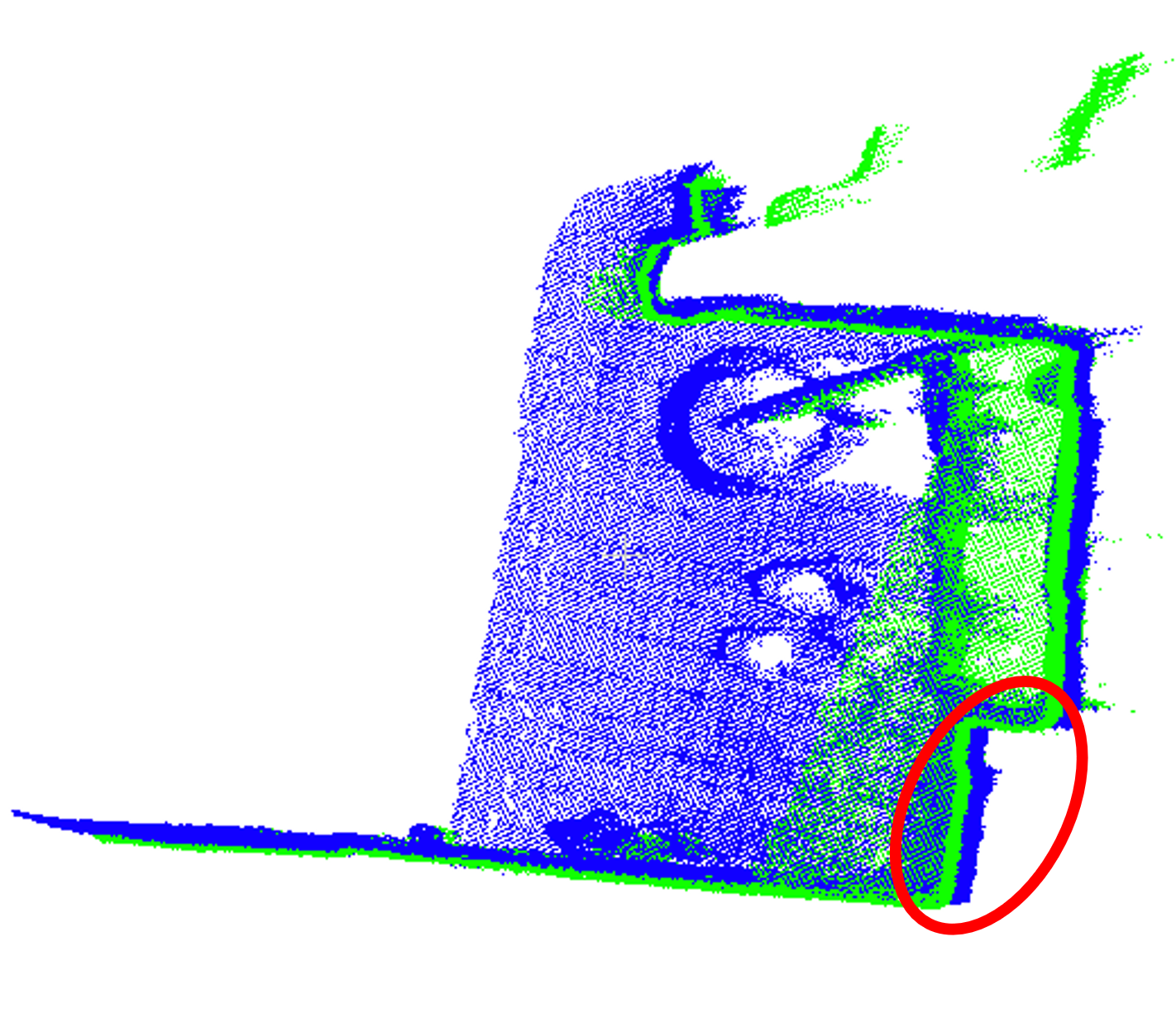} &
\vspace{0.1cm}\includegraphics[width=0.12\textwidth]{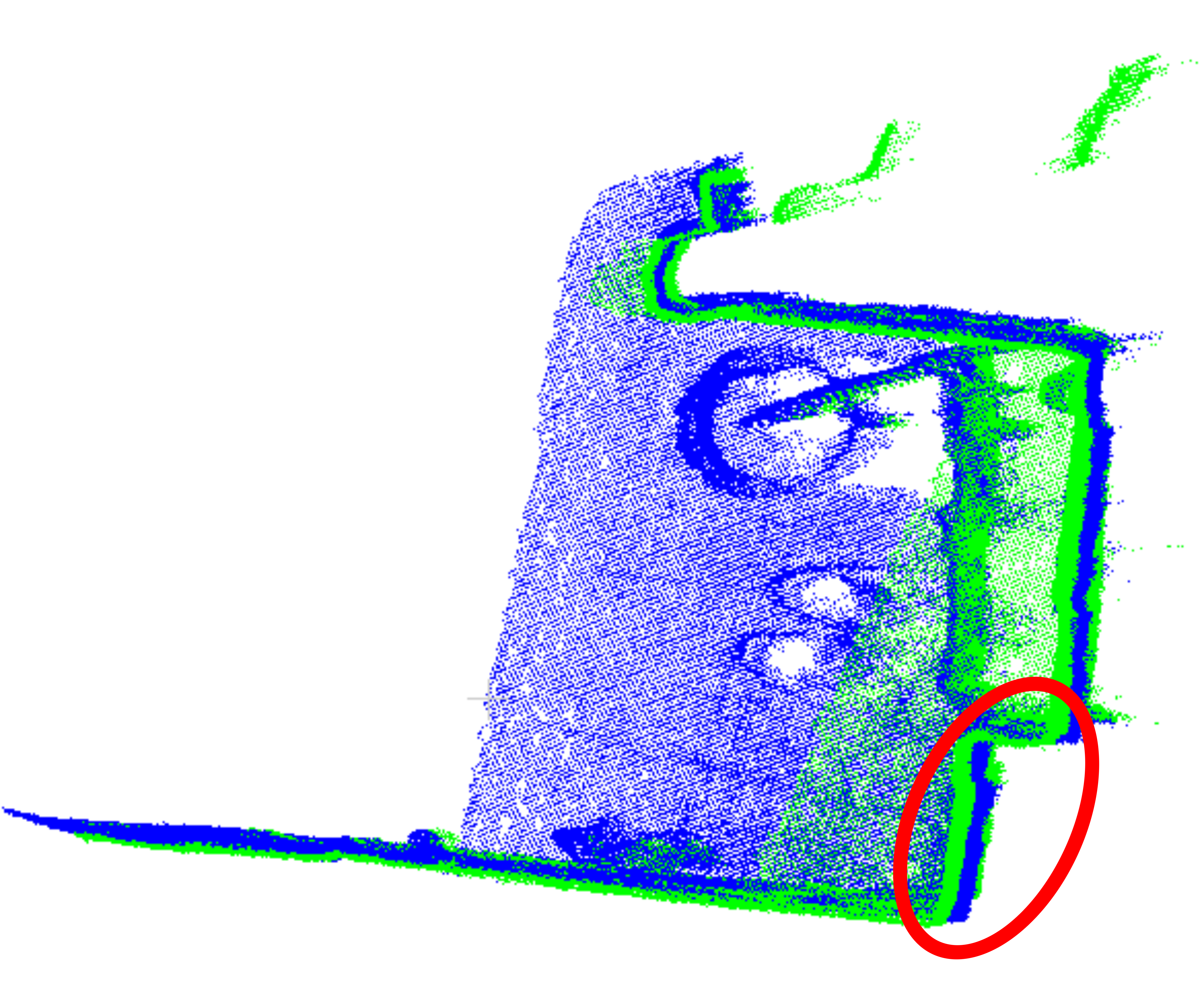} & 
\vspace{0.1cm}\includegraphics[width=0.12\textwidth]{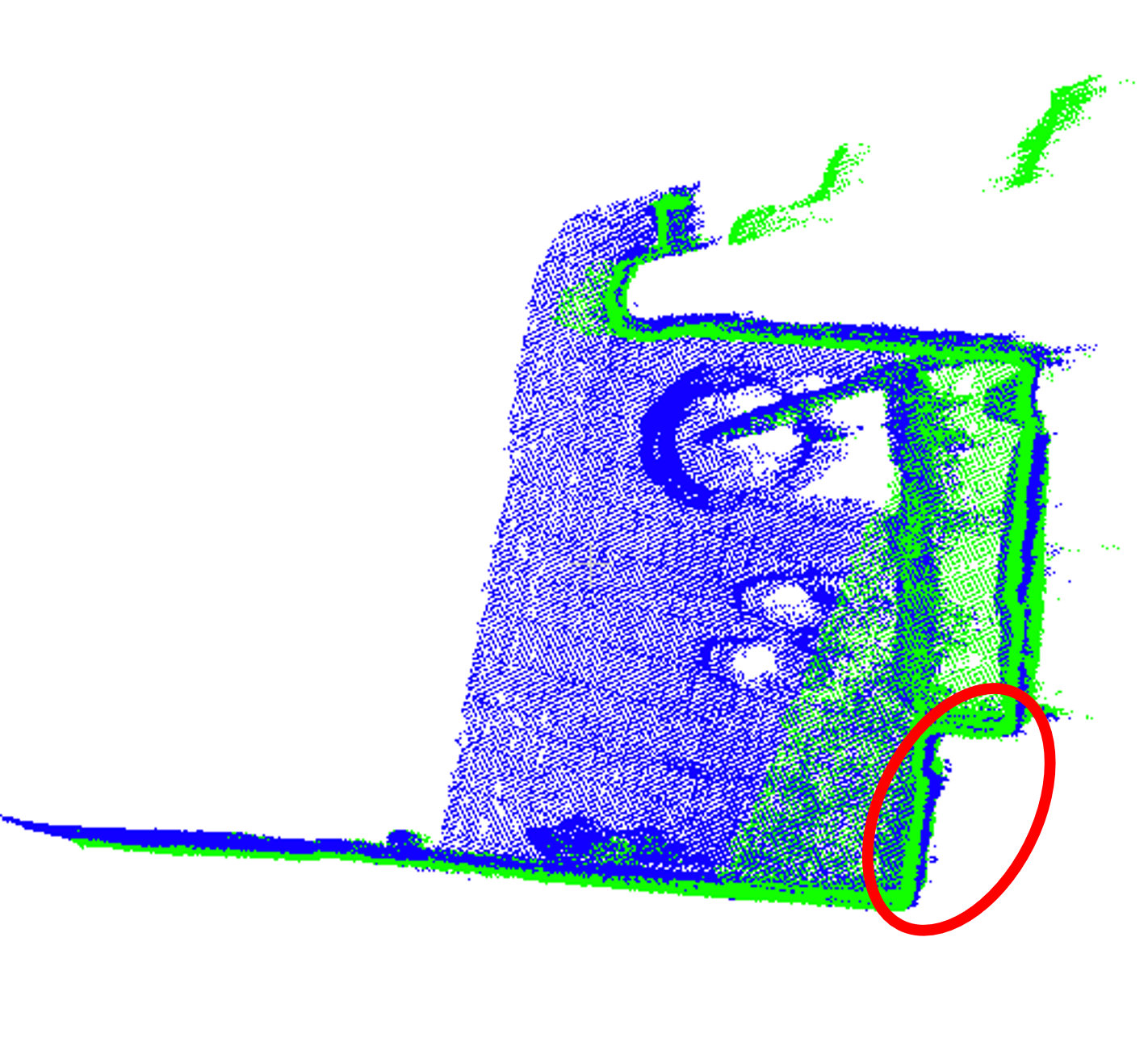} \\[0.2cm] 

\vspace{0.2cm}
3DMatch\_mit\_lab 
\vspace{0.2cm} & 
\includegraphics[width=0.12\textwidth]{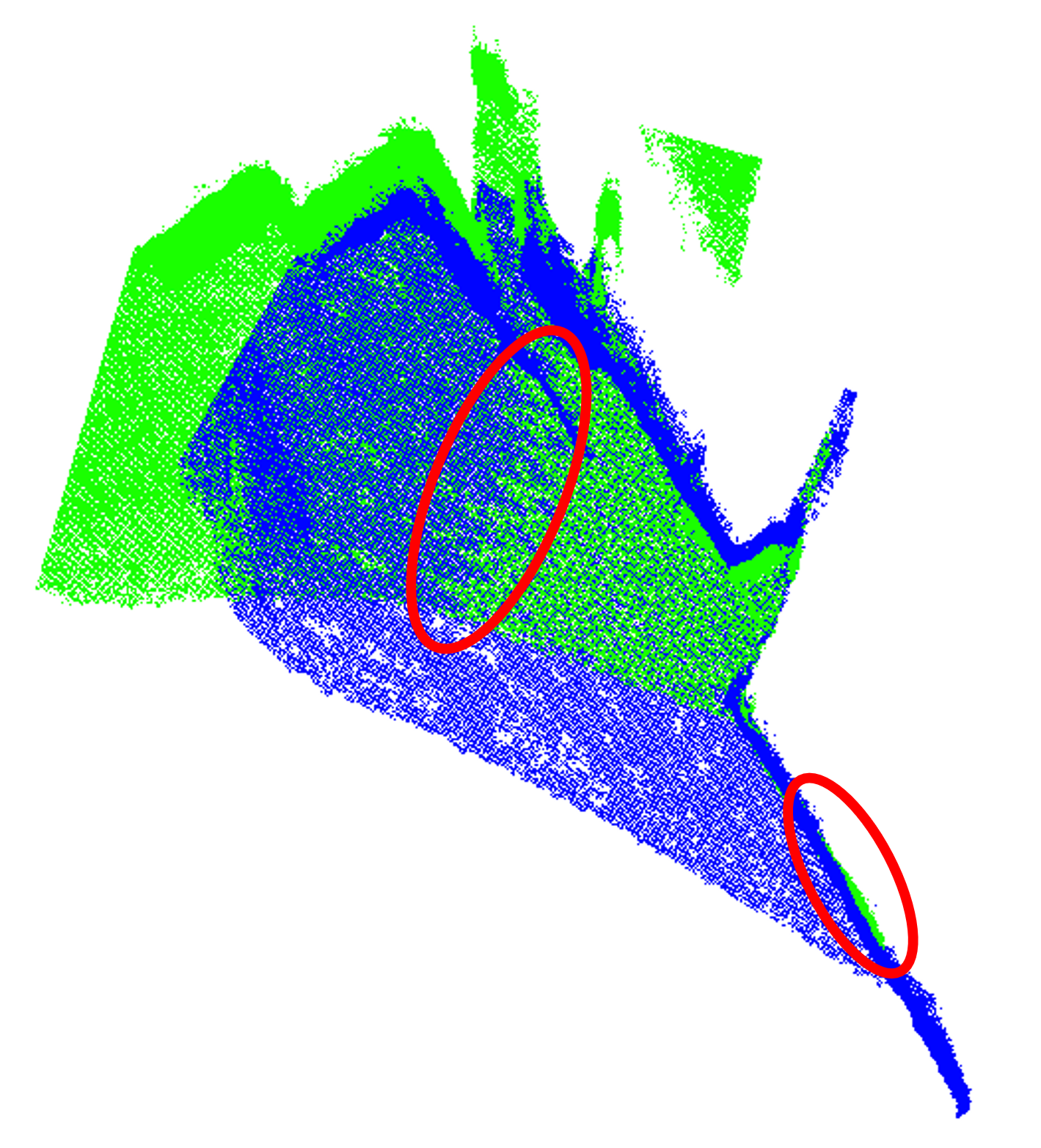} &
\includegraphics[width=0.12\textwidth]{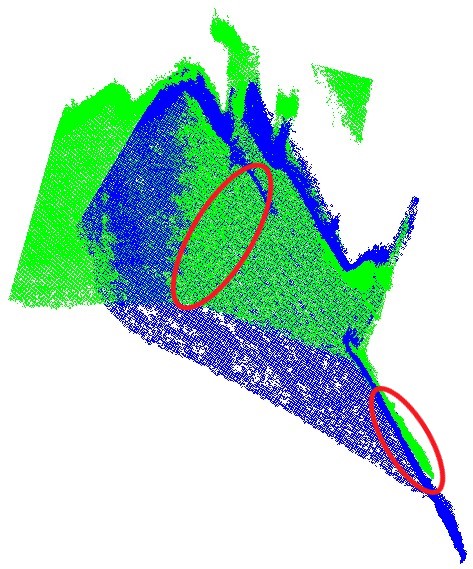} &
\includegraphics[width=0.12\textwidth]{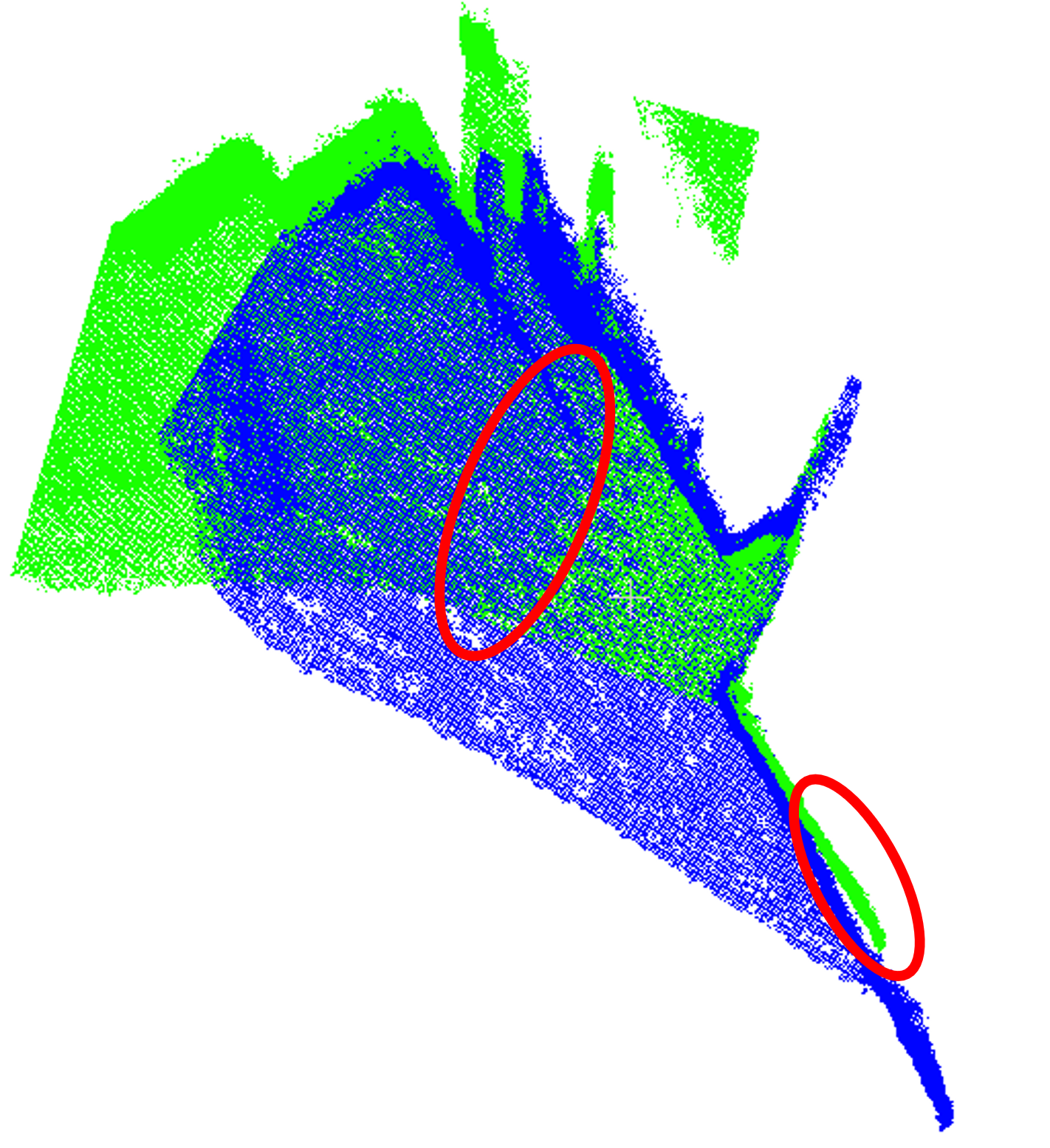} &
\includegraphics[width=0.12\textwidth]{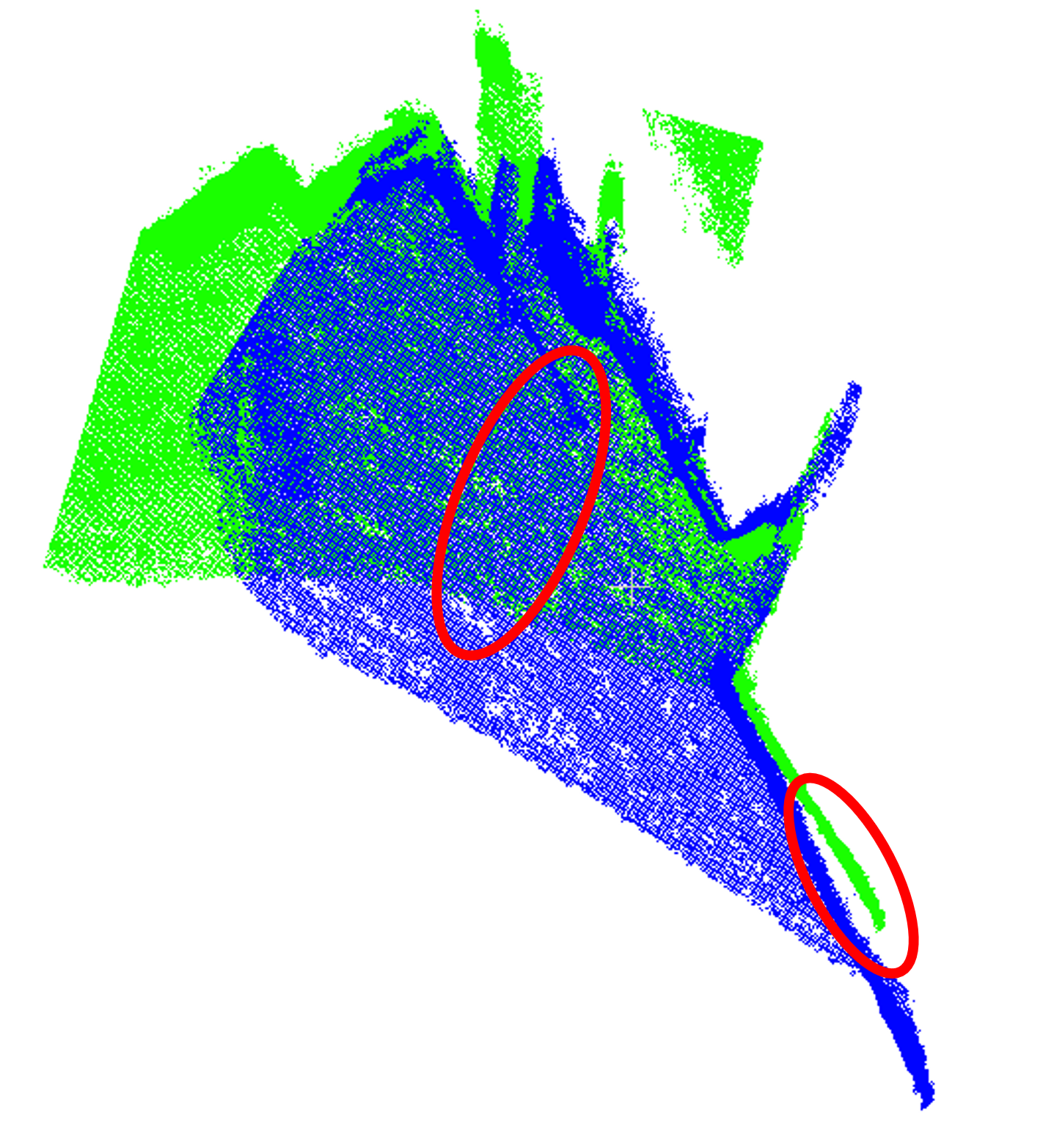} &
\includegraphics[width=0.12\textwidth]{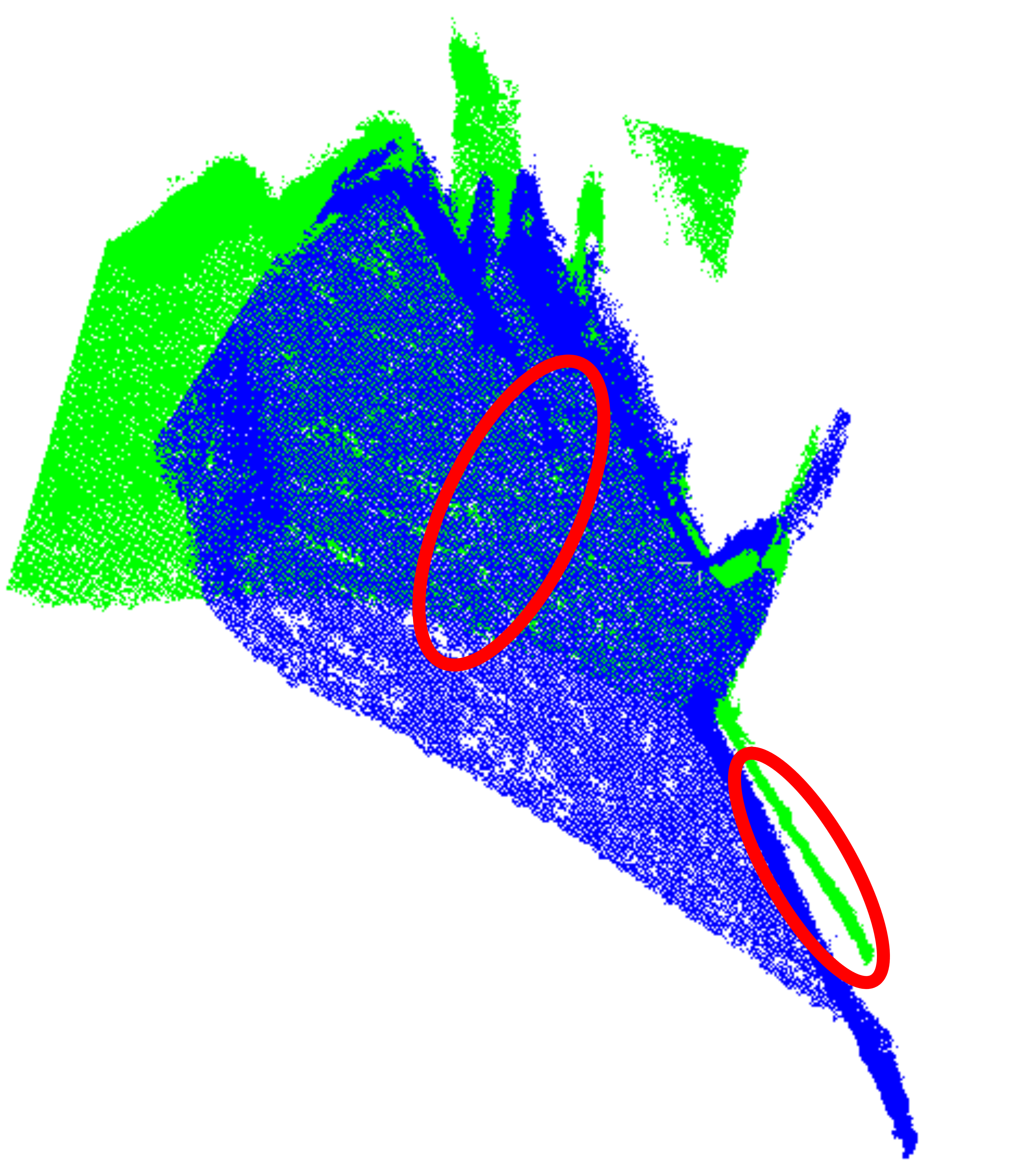} & 
\includegraphics[width=0.12\textwidth]{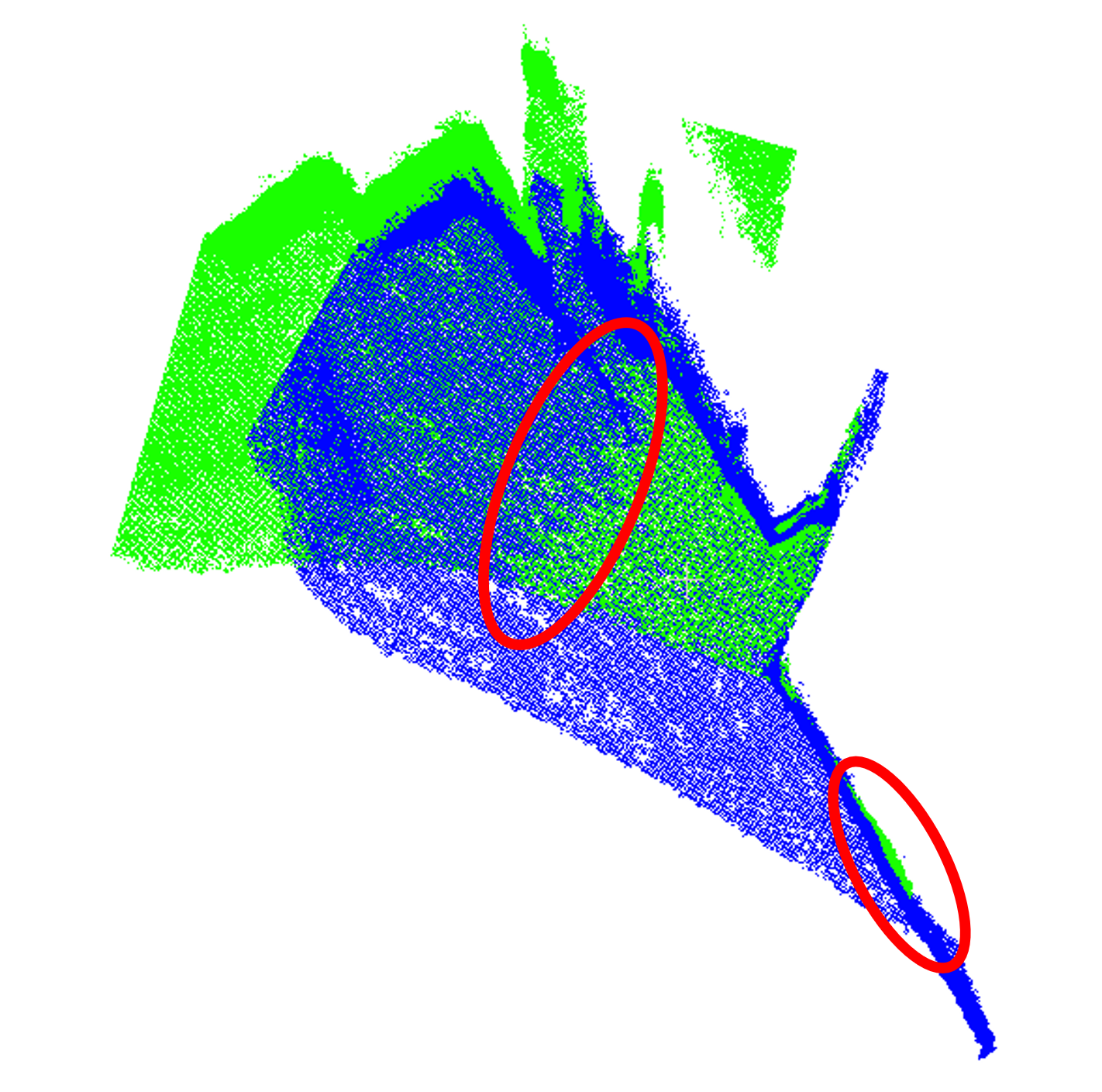} \\[0.2cm]

\vspace{0.2cm}
KITTI\_43
\vspace{0.2cm} & 
\includegraphics[width=0.12\textwidth]{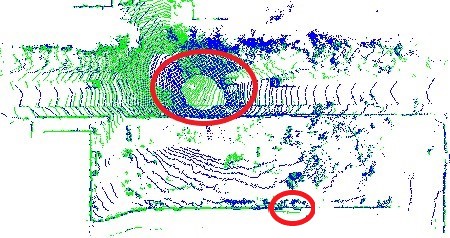} &
\includegraphics[width=0.12\textwidth]{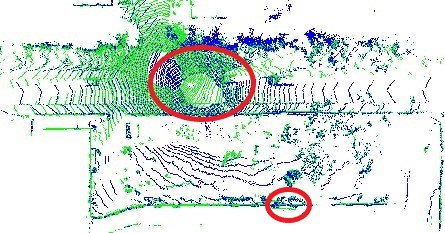} &
\includegraphics[width=0.12\textwidth]{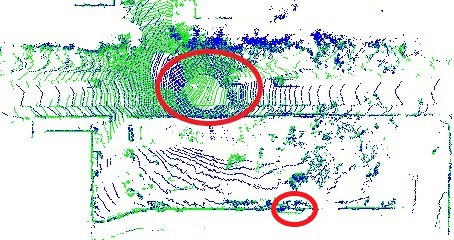} &
\includegraphics[width=0.12\textwidth]{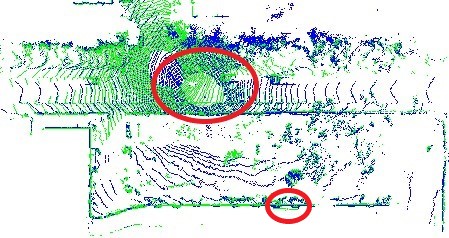} &
\includegraphics[width=0.12\textwidth]{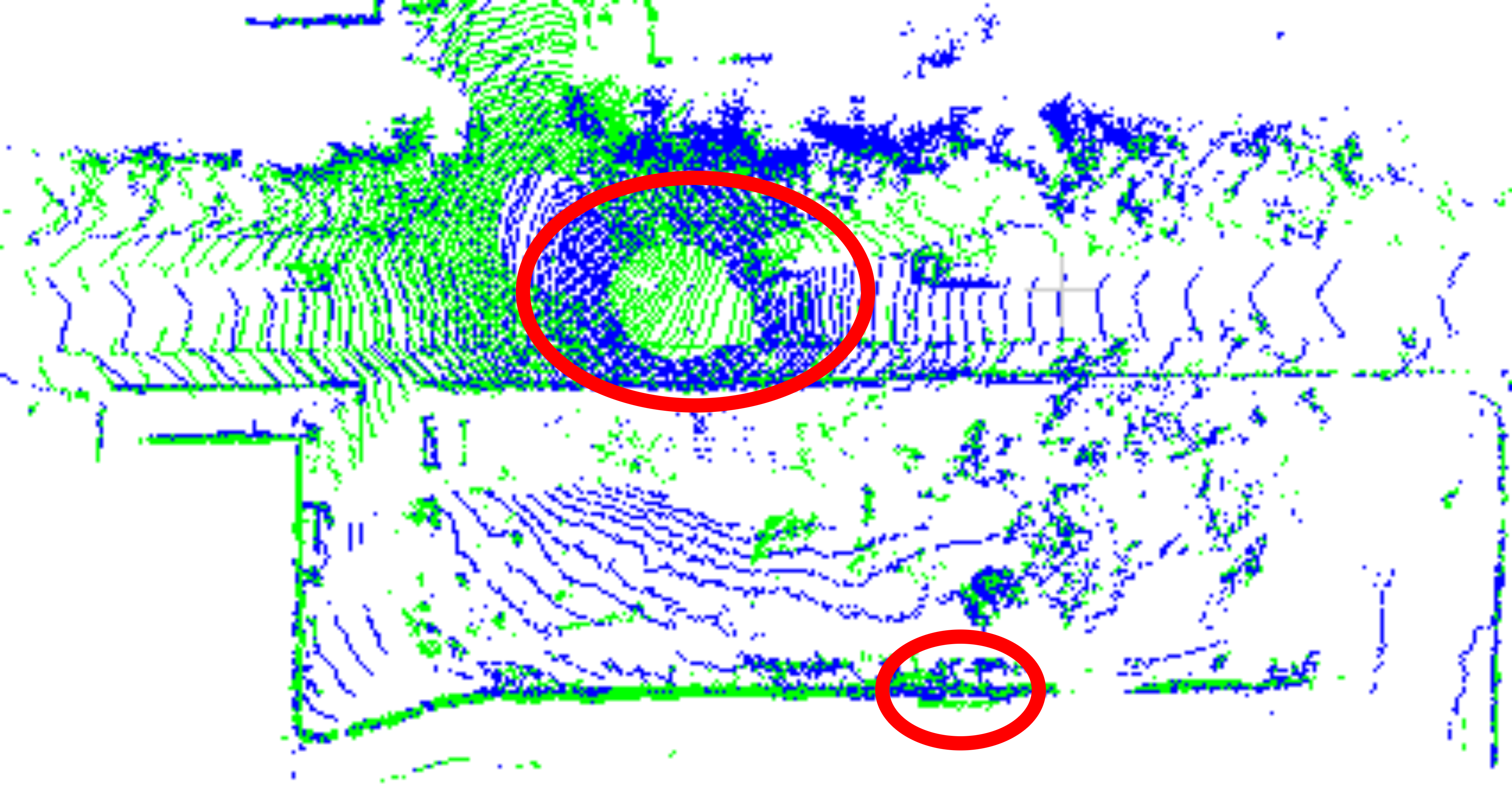} & 
\includegraphics[width=0.12\textwidth]{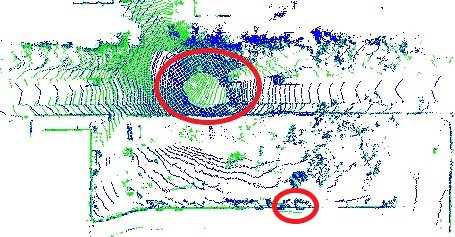} \\[0.2cm]

\vspace{0.2cm}
KITTI\_330
\vspace{0.2cm} & 
\includegraphics[width=0.12\textwidth]{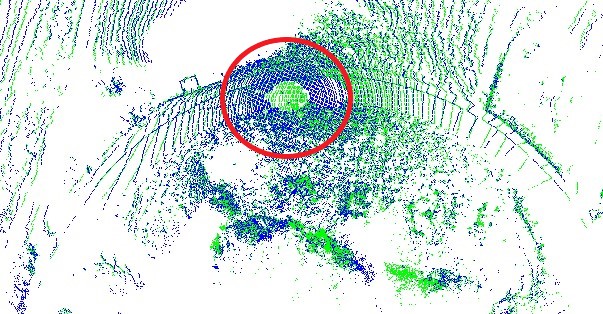} &
\includegraphics[width=0.12\textwidth]{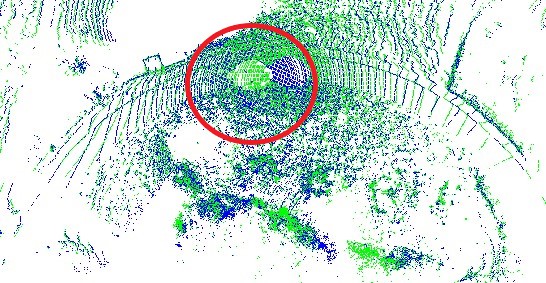} &
\includegraphics[width=0.12\textwidth]{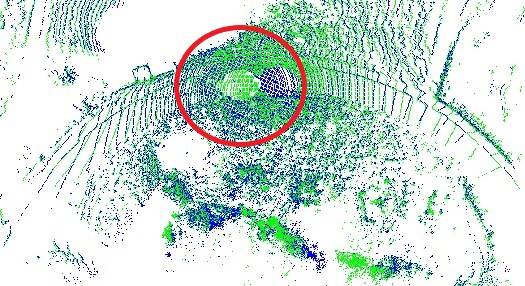} &
\includegraphics[width=0.12\textwidth]{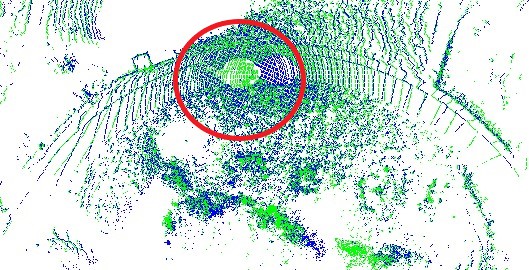} &
\includegraphics[width=0.12\textwidth]{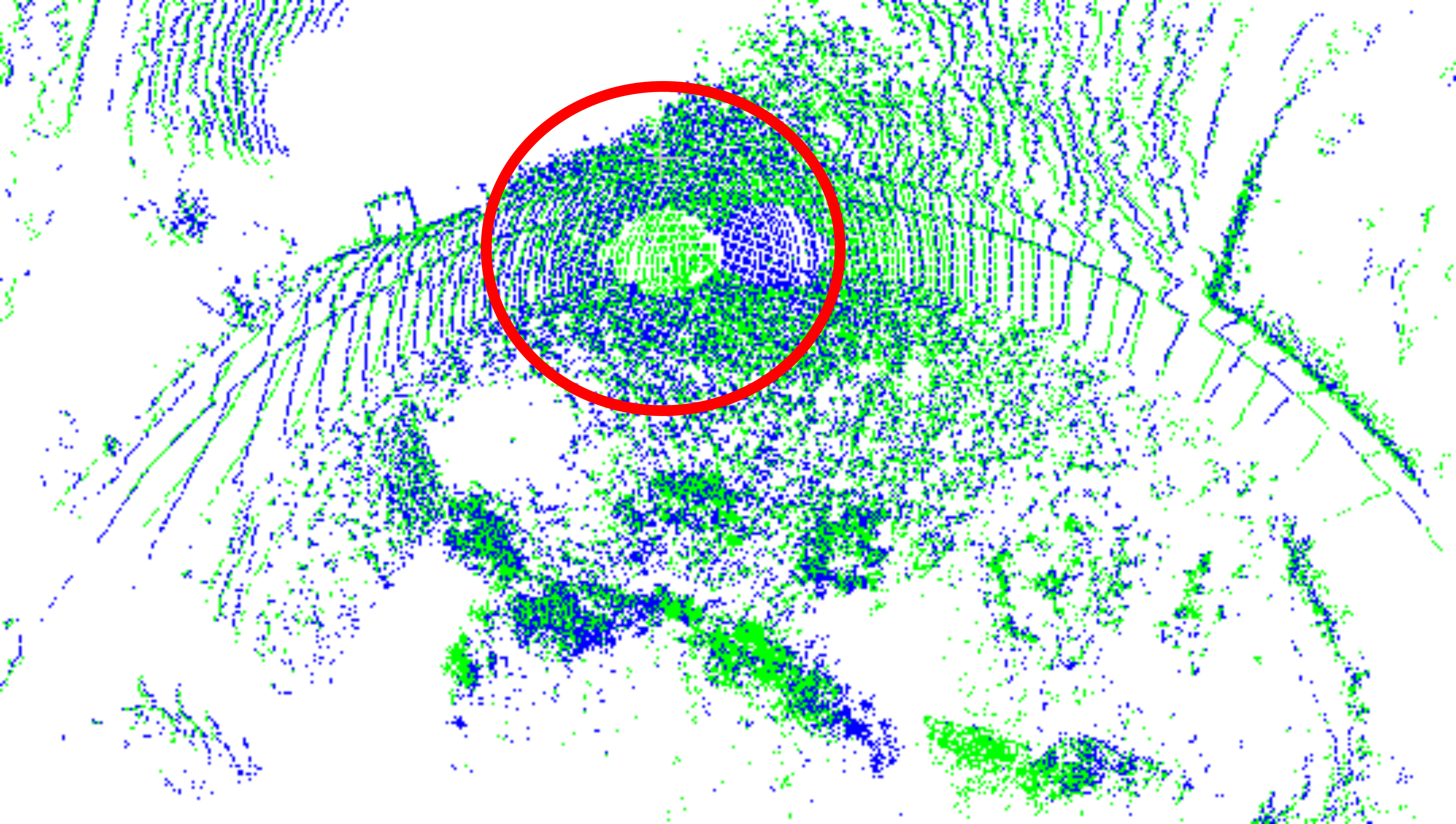} & 
\includegraphics[width=0.12\textwidth]{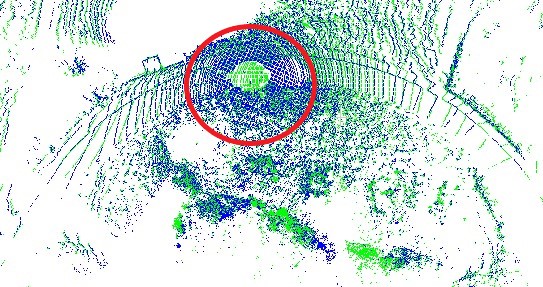} \\
[0.2cm]

\vspace{0.2cm}
WHU-TLS\_park\_1 
\vspace{0.2cm} & 
\includegraphics[width=0.12\textwidth]{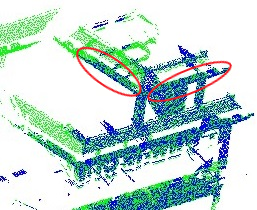} &
\includegraphics[width=0.12\textwidth]{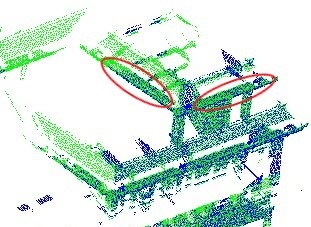} &
\includegraphics[width=0.12\textwidth]{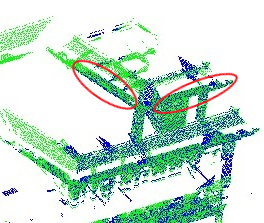} &
\includegraphics[width=0.12\textwidth]{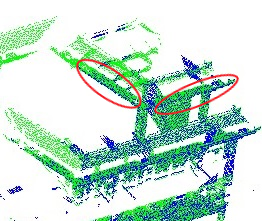} &
\includegraphics[width=0.12\textwidth]{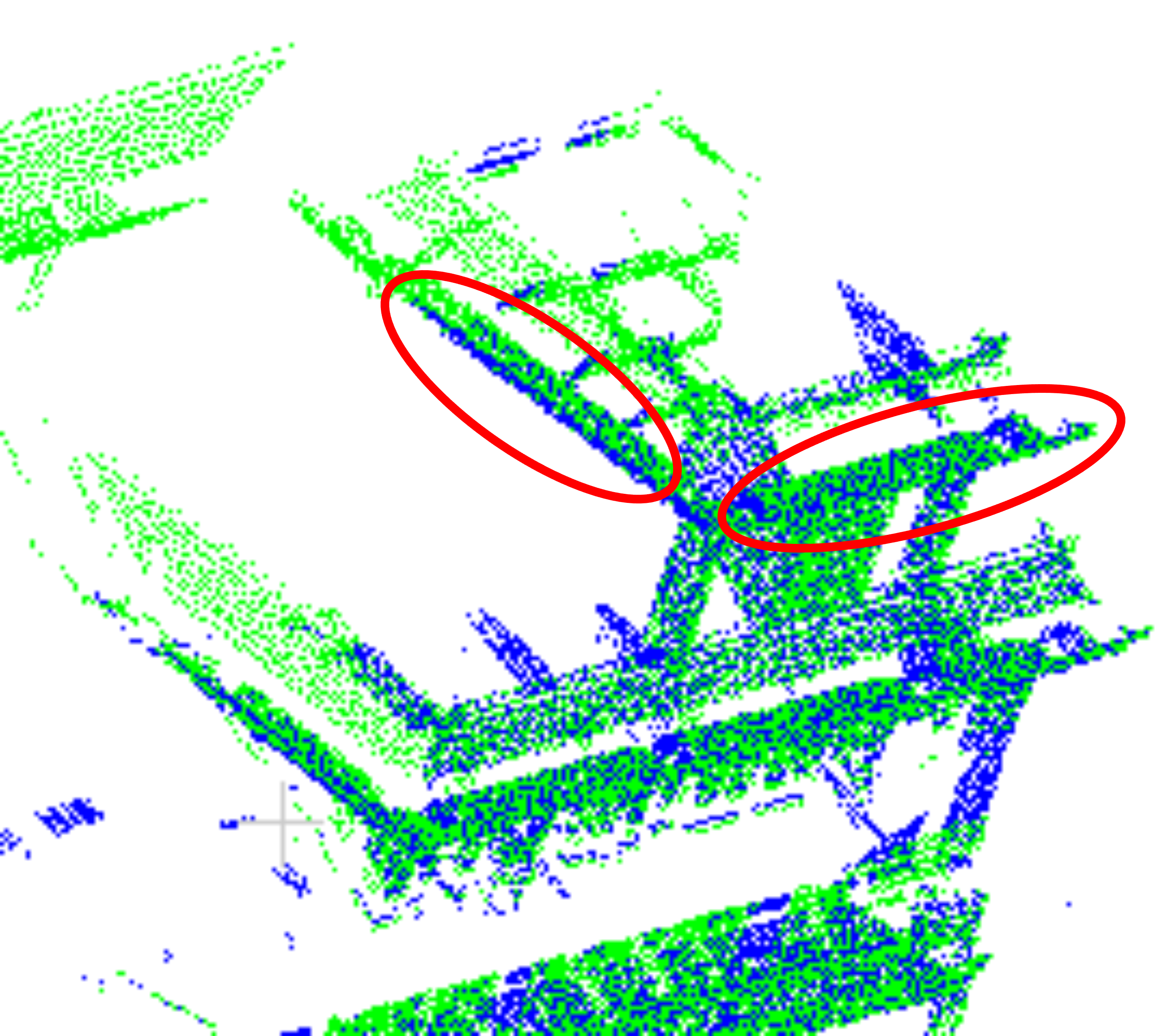} & 
\includegraphics[width=0.12\textwidth]{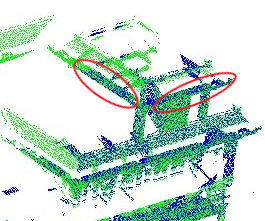} \\
[0.2cm]

\vspace{0.2cm}
WHU-TLS\_park\_2 
\vspace{0.2cm} & 
\includegraphics[width=0.12\textwidth]{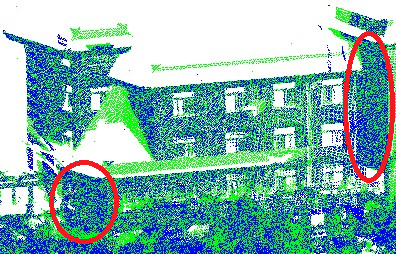} &
\includegraphics[width=0.12\textwidth]{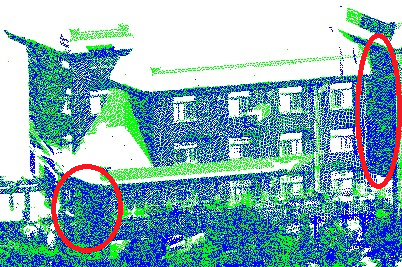} &
\includegraphics[width=0.12\textwidth]{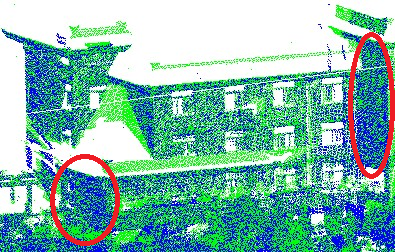} &
\includegraphics[width=0.12\textwidth]{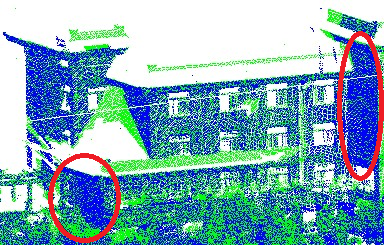} &
\includegraphics[width=0.12\textwidth]{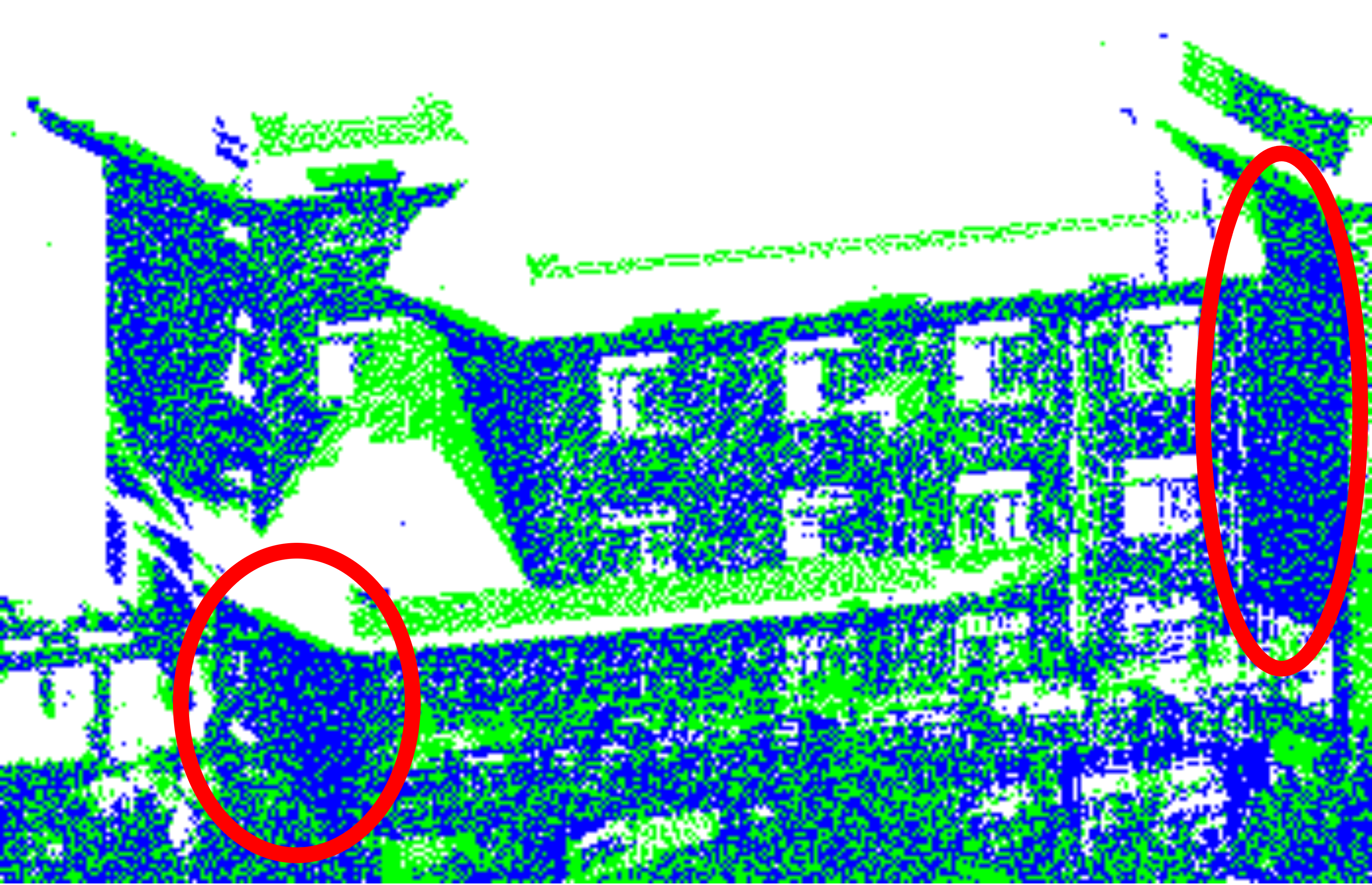} & 
\includegraphics[width=0.12\textwidth]{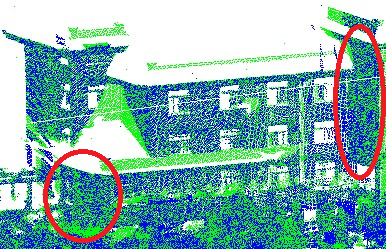} \\
[0.2cm] \bottomrule
\end{tabular}
\end{table*}

\begin{table*}[htbp] 
    \captionsetup{justification=raggedright, singlelinecheck=false, labelsep=newline}
    \centering
    \caption{The ablation study of our PCR algorithm.} 
    \label{tab:ablation}
    
    \small 
    
    \begin{tabular*}{\textwidth}{@{\extracolsep{\fill}} cc|cccccccc @{}} \toprule
        AHS-LVLP & SUS & ${\rm{R}}^{err}$ (\textdegree) & $T^{err}$ (cm) & RMSE & MeSE & P (\%) & R (\%) & F1 (\%) & Time (s) \\ \midrule
        {\ding{53}} & {\ding{53}} & 0.834 & 8.736 & 0.120 & 0.097 & 93.893 & 94.786 & 94.269 & 1.524 \\
        {\ding{53}} & {\ding{51}} & 0.821 & 7.114 & 0.107 & 0.085 & 93.752 & 95.555 & 94.580 & 1.544 \\
        {\ding{51}} & {\ding{53}} & 0.864 & 9.230 & 0.133 & 0.106 & 93.685 & 93.610 & 93.545 & 0.119 \\
        {\ding{51}} & {\ding{51}} & 0.823 & 6.575 & 0.083 & 0.069 & 95.072 & 95.674 & 95.304 & 0.121 \\ \bottomrule
    \end{tabular*} 
\end{table*}

\subsubsection{Visual quality comparison and the ablation study of our algorithm}
From the quantitative quality comparison discussed in the last subsection, it is known that the four methods, namely our algorithm, TEAR, $\text{SC}^2$-PCR++, and One-point RANSAC, rank top four. Therefore, the visual quality comparison of the four methods is presented. Secondly, in terms of the global early termination and SUS, the ablation study of our algorithm is reported.

\vspace{1ex}
\noindent \quad a) Visual quality comparison:
As listed in the first column of Table~\ref{tab:showcase}, six testing point cloud pairs, namely 3DMatch\_home\_md and 3DMatch\_mit\_lab; KITTI\_43 and KITTI\_330; WHU-TLS\_park\_1 and WHU-TLS\_park\_2, are used to show the perceptual merit of our algorithm relative to the One-point RANSAC, $\text{SC}^2$-PCR++, TEAR, and MAC++ methods. The ground truth alignment results for the six testing point cloud pairs are demonstrated in the second column of Table~\ref{tab:showcase}, where the aligned source point cloud is marked in blue and the target point cloud is marked in green. After performing the five considered methods on the six testing point cloud pairs, the third, fourth, fifth, sixth, and seventh columns demonstrate their perceptual results.

The first and second rows of Table~\ref{tab:showcase} demonstrate the visual results of the five considered methods for the point cloud pairs,  
``3DMatch\_home\_md'' and ``3DMatch\_mit\_lab'', respectively. Taking the ground truth aligned point cloud pair as the baseline, it is observed that in the areas within the red ellipses, One-point RANSAC and our algorithm achieve the best alignment results for the first and second point cloud pairs, respectively.

The third and fourth rows of Table~\ref{tab:showcase} demonstrate the visual results of all considered methods for the point cloud pairs ``KITTI\_43'' and ``KITTI\_330'', respectively. Using the two ground truth aligned point cloud pairs as references, we observe that within the regions enclosed by the red ellipses, our algorithm achieves the best alignment on both pairs, whereas the four comparative methods exhibit noticeable misalignments.

The fifth and sixth rows of Table~\ref{tab:showcase} demonstrate the visual results of all considered methods for the WHU-TLS point cloud pairs ``WHU-TLS\_park\_1'' and ``WHU-TLS\_park\_2'', respectively. As can be seen in the red ellipse regions, our algorithm yields the most precise and complete alignments, closely matching the ground truth. In contrast, the four comparative methods exhibit slight misalignments or residual offsets, manifested as visible blue-green discrepancies in these regions. The superior alignment of our method in these large-scale terrestrial laser scanning (TLS) scenes further confirms its robustness under extremely high outlier ratios and uneven point densities, which are typical characteristics of the WHU-TLS dataset.

\vspace{1ex}
\noindent \quad b) Ablation study:
\label{sec:ablation study}
To validate the effectiveness of the proposed algorithm, the ablation study evaluates two primary components: the AHS and LVLP methods, simply called AHS-LVLP, and the SUS. Table~\ref{tab:ablation} presents the registration metrics under four different configurations of these two components. Notably, the baseline configuration, where both components are disabled, corresponds to the centralized RANSAC (C-RANSAC) method \citep{wt2024cransac}.

In the first configuration (baseline), it yields ${\rm R}^{err} = 0.834^{\circ}$, ${T}^{err} = 8.736$~cm, F1 = $94.269\%$, and Time = $1.524$~s. When only the SUS is enabled (the second configuration), the local correspondence set is dynamically updated during the iterations. This component decreases ${\rm R}^{err}$ to $0.821^{\circ}$ and ${T}^{err}$ to $7.114$~cm, while the recall rate increases to $95.555\%$ and the F1 score increases to $94.580\%$. The execution time changes from $1.524$~s to $1.544$~s, indicating that the continuous updating process increases a little runtime cost.

In the third configuration, only the AHS-LVLP component is enabled. Compared to the baseline, the execution time decreases from $1.524$~s to $0.119$~s. However, the ${\rm R}^{err}$ increases to $0.864^{\circ}$ and the ${T}^{err}$ increases to $9.230$~cm, while the F1 score decreases to $93.545\%$. This behavior indicates that the AHS-LVLP component reduces the correspondence search space to accelerate the estimation process, but filtering correspondences without the continuous refinement from SUS excludes some valid pairs, causing numerical fluctuation in the error metrics.

In the fourth configuration, both AHS-LVLP and SUS are enabled simultaneously. Our algorithm achieves ${\rm R}^{err} = 0.823^{\circ}$ and ${T}^{err} = 6.575$~cm. The precision and recall rates increase to $95.072\%$ and $95.674\%$, respectively, resulting in an F1 score of $95.304\%$. The execution time is $0.121$~s. These values confirm that executing both AHS-LVLP and SUS simultaneously provides the lowest translation error and highest accuracy metrics among the four configurations, achieving clear registration accuracy improvement and runtime reduction relative to the baseline method.
\section{Conclusions}\label{sec:conclusions}
The proposed probabilistic self-update (PSU) correspondence and line vector set-based PCR algorithm has been presented. In our algorithm, the proposed AHS-based and LVLP-based methods can construct the initial local correspondence and line vector sets quickly, although the initially constructed two local sets contain more outliers. Fortunately, after one RANSAC interaction round, the proposed PSU strategy can update the two local sets to increase the compatibility of the two local sets. Ablation study has verified the accuracy and time merits of the proposed PSU strategy. As for the proposed global termination condition, the ablation study also has verified the time merit, although it may sacrifice accuracy slightly. Based on three public datasets, rigorous experimental data have demonstrated that our algorithm delivers clear accuracy and time improvements relative to six state-of-the-art methods. The future work is to strengthen the results of this paper to enhance the accuracy for low overlapping cases.

\section*{CRediT authorship contribution statement}
\textbf{Kuo-Liang Chung:} Conceptualization, Methodology, Formal analysis, Writing - Original Draft. \textbf{Yu-Cheng Lin:} Software, Validation, Writing - Review \& Editing, Data Curation. \textbf{Wu-Chi Chen:} Software, Validation, Writing - Review \& Editing.

\section*{Declaration of competing interest}
The authors declare that they have no known competing financial interests or personal relationships that could have appeared to influence the work reported in this paper.

\section*{Acknowledgments}
This work was supported under the Grant MOST-114-2221-E-011-040-MY3 of the Ministry of Science and Technology, Taiwan.

\section*{Data availability}
The datasets (3DMatch, KITTI, and WHU-TLS) used during the current study are publicly available. The C++ source code of the proposed algorithm is openly available at \href{https://github.com/ivpml84079/Probabilistic-Self-Update-Line-Vector-Set-Based-Point-Cloud-Registration}{https://github.com/ivpml84079/Probabilistic-Self-Update-Line-Vector-Set-Based-Point-Cloud-Registration}. Additional data that support the findings of this study are available from the corresponding author upon reasonable request.

\bibliographystyle{elsarticle-num}

\setlength{\bibsep}{0pt plus 0.3ex} 

\renewcommand{\bibfont}{\footnotesize} 

{
    \begin{spacing}{1.0}
        \bibliography{ref} 
    \end{spacing}
}

\end{document}